\documentclass{article}

\usepackage{microtype}
\usepackage{graphicx}
\usepackage{subfigure}
\usepackage{booktabs} 
\usepackage{listings}
\usepackage{tabularx}
\usepackage{hyperref}

\usepackage[accepted]{icml2024}

\usepackage{amsmath}
\usepackage{amssymb}
\usepackage{mathtools}
\usepackage{amsthm}
\usepackage{svg}
\usepackage{float}

\usepackage[capitalize,noabbrev]{cleveref}

\theoremstyle{plain}

\theoremstyle{definition}

\theoremstyle{remark}

\usepackage[textsize=tiny]{todonotes}

\newcommand{\learnedR}{\ensuremath{R_{\theta}}}
\newcommand{\state}{\ensuremath{\mathcal{S}}}
\newcommand{\action}{\ensuremath{\mathcal{A}}}
\newcommand{\reward}{\ensuremath{\mathcal{R}}}
\newcommand{\transition}{\ensuremath{\mathcal{T}}}
\newcommand{\horizon}{\ensuremath{T}}
\newcommand{\traj}[1]{\ensuremath{\tau}_{#1}}
\newcommand{\dataset}{\ensuremath{\mathcal{D}}}

\newcommand{\replay}{\ensuremath{\mathcal{D}_{RB}}}
\newcommand{\rank}[1]{\ensuremath{Rank_{#1}}}

\begin{document}

\twocolumn[
\icmltitle{Learning Reward for Robot Skills Using Large Language Models via Self-Alignment}



\icmlsetsymbol{equal}{*}

\begin{icmlauthorlist}
\icmlauthor{Yuwei Zeng}{sch1}
\icmlauthor{Yao Mu}{sch2,sch1}
\icmlauthor{Lin Shao}{sch1}
\end{icmlauthorlist}

\centering{
    \texttt{zyuwei@comp.nus.edu.sg, muyao@connect.hku.hk, linshao@nus.edu.sg}
}

\icmlaffiliation{sch1}{National University of Singapore}
\icmlaffiliation{sch2}{The University of Hong Kong}

\icmlcorrespondingauthor{Lin Shao}{linshao@nus.edu.sg}

\icmlkeywords{Machine Learning, ICML}

\vskip 0.3in
]



\printAffiliationsAndNotice{}  

\begin{abstract}
Learning reward functions remains the bottleneck to equip a robot with a broad repertoire of skills. Large Language Models (LLM) contain valuable task-related knowledge that can potentially aid in the learning of reward functions. However, the proposed reward function can be imprecise, thus ineffective which requires to be further grounded with environment information. We proposed a method to learn rewards more efficiently in the absence of humans. Our approach consists of two components: We first use the LLM to propose features and parameterization of the reward, then update the parameters through an iterative self-alignment process. In particular, the process minimizes the ranking inconsistency between the LLM and the learnt reward functions based on the execution feedback. The method was validated on 9 tasks across 2 simulation environments. It demonstrates a consistent improvement over training efficacy and efficiency, meanwhile consuming significantly fewer GPT tokens compared to the alternative mutation-based method. Project website: \href{https://sites.google.com/view/rewardselfalign}{https://sites.google.com/view/rewardselfalign}.
\end{abstract}
\begin{figure*}[t!]
\centering
    \includegraphics[width=\textwidth]{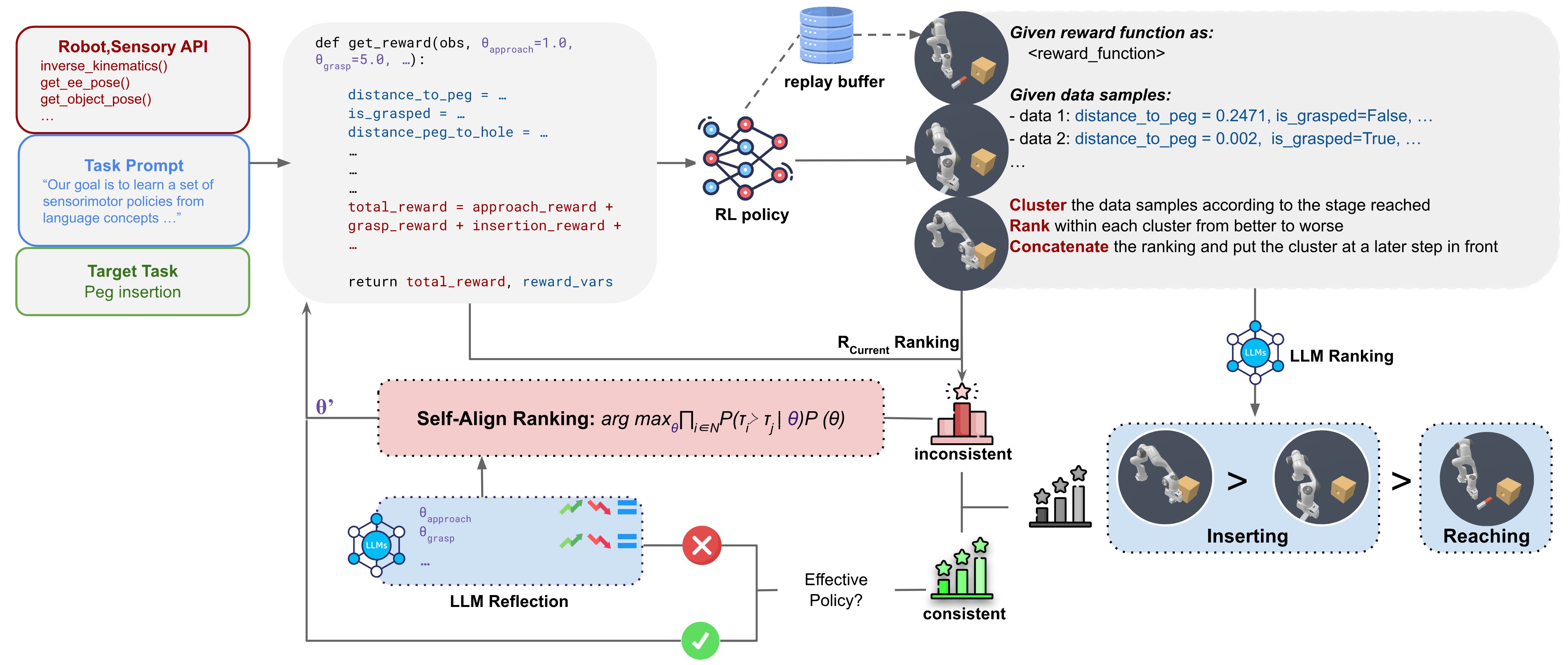}
    \vspace{-20pt}
\caption{\small The overview of our method. We learn the reward function using LLM with a bi-level optimization structure. We first use the LLM to propose features and parameterization of the reward function. Next, we update the parameters of this proposed reward function through an iterative self-alignment process. In particular, this process minimizes the ranking inconsistency between the LLM and our learned reward functions based on the new observations.}
\vspace{-15pt}
\label{fig:para_proposal}
\end{figure*}

\vspace{-20pt}
\section{Introduction}
Reinforcement learning has demonstrated the effectiveness in acquiring complex skills from walking over uneven terrain \cite{valsecchi2020quadrupedal, manchester2011stable} to dexterous manipulation\cite{akkaya2019solving, chen2023visual}. However, such effectiveness largely depends on a well-designed reward function that relies on expert knowledge of tasks, followed by non-trivial tuning often to both optimize the efficacy and prevent the policy from exploiting flaws that can be easily introduced during reward shaping. Inverse reinforcement learning (IRL) \cite{abbeel2004apprenticeship,ho2016generative} seeks to resolve this issue by automatically learning the reward function from expert demonstrations. Still the process can be exceedingly costly for expert demonstration gathering to cover the vast variety and complexity of the state space and yield a robust control. 

Recently, Large Languaged Models (LLMs), trained using extensive human data, have demonstrated to be embedded with richly useful task-related knowledge. Several existing studies have used LLM to directly propose action \cite{liang2023code} or reward values \cite{kwon2023reward, adeniji2023language}. However, using LLM to learn reward functions still presents a challenge due to the task sensitivity to the exact numerical values while LLM shows limited capacity. In addition, such value setting in general requires to be grounded to the specific setup. For example, In \cite{yu2023language} LLM set the target torso height for a quadruped to 0.3m for moon walk instruction. With further inquiring about rationale behind the number, LLM explains "the specific value here is arbitrary and should be adjusted based on the robot's design and requirements". Previous works address such limitation with human feedback or in-context example \cite{yu2023language, xie2023text2reward, wang2023robogen}, or evolutionary method \cite{ma2023eureka} with success rate feedback to LLM, both are still in a form of trial and error from LLM update, that might end up with fluctuated instead of improving performance.

In this work, we aim to answer the question: is there a way we can learn the reward more efficiently in the absence of humans? It is observed that LLMs have shown promising ability in summarizing and classifying text, which allows them to effectively distinguish different observations in textual form. We propose to utilize such ability to extract ranking signals from LLM, which could be more robust to guide reward learning than direct value prediction of the parameters. We first utilize LLM to break down a task into steps with Dos and Don$'$ts through Chain of Thought (CoT) \cite{wei2022chain} and propose the initial reward parameterization, particularly the feature selection and template structure. Next, we iteratively update the parameters of the proposed reward function in a self-alignment process which operates on a double-loop structure. The inner loop induces the optimal policy from the current reward function, samples trajectories using this policy and generates execution descriptions with the proposed reward features. The outer loop updates the reward parameters by aligning the ranking between LLM proposed with the execution description feedback, and the ranking from the current reward function. When no discrepancy exists yet no effective policy is developed, we also actively adjust reward parametrization in the direction LLM reflection hints \cite{liu2023reflect}, and numerically optimize it to keep the same ranking self-consistency. This process is similar to IRL's bi-level optimization structure, with one key difference in the outer loop: instead of minimizing differences between expert demonstrations \cite{ke2021imitation}, our method employs ranking from LLM. Since all supervision signals come from LLM, we describe this as the self-alignment reward update.

To summarize, out contribution includes:
\vspace{-5pt}
\begin{itemize}
    \vspace{-5pt}
    \item We proposed a framework to learn the reward functions with LLM through an interative self-alignment process, which periodically updates the reward function to minimize the ranking inconsistency of execution generated from LLM and the current reward function. 
    \vspace{-5pt}
    \item Leveraged upon the self-alignment process, we included active parameter adjustment with LLM heuristic to improve reward saliency, while preventing it from unintentional flaw through enforcing the consistency.
    \vspace{-5pt}
    \item We validated the framework on 9 tasks on 2 simulation environments. It demonstrates a consistent improvement over training efficacy and efficiency while being token efficient compared to alternative method.
\vspace{-5pt}
\end{itemize}

\section{Related Work}

\paragraph{Inverse Reinforcement Learning from Human Preference} 
Inverse reinforcement learning (IRL) studies how to autonomously learn a reward function from expert data. Traditional IRL methods utilise the optimal expert demonstrations as supervision signals, aiming to learn a reward function that can encode an objective consistent with the expert demonstrations\cite{ng2000algorithms, abbeel2004apprenticeship, ho2016generative, hadfield2017inverse}. The bottleneck of IRL is the quantity and quality of the expert demonstrations. Some other works explore other forms of expert supervision, such as trajectory preference/ranking \cite{sadigh2017active, palan2019learning, lee2021pebble, biyik2022learning, mehta2022unified}. Learning from preference ranking offers some benefits over standard IRL, which relies on almost perfect expert demonstrations for imitation. Essentially, it learns from ranked sub-optimal demonstrations, potentially surpassing the original demonstrator's performance. This makes it less demanding for experts to create only optimal behaviours and also encodes further contrasting information about preferable and undesirable behaviours. A major downside, however, remains the need to amass this information. We investigate the potential to extract task-related data from a pre-trained large language model, effectively bypassing this data collection challenge.

\vspace{-10pt}
\paragraph{LLM for robotics reward learning}
Large Language Models (LLMs) have shown great potential as cost-effective tools for extracting task objectives as they embed substantial human knowledge \cite{huang2022language, brohan2023can}. A few works proposed to generate code with large language models such as ProgPrompt \cite{singh2023progprompt}, Policy-as-Code \cite{liang2023code}. Based on this, several prior studies have further utilized LLMs to predict reward functions. \cite{kwon2023reward} learns to translate nature language instructions to reward values, \cite{yu2023language, xie2023text2reward, wang2023robogen} proposed robot skill synthesis or task learning with LLM-generated reward function and further refine with human feedback or in-context example. The work either uses LLMs in a zero-shot manner implying their reward function can't utilize environmental feedback, or the update relies on human intervention which can be laborious. \cite{ma2023eureka} proposed to automatically generate and select the reward function through an evolutionary manner which achieved human-level reward design. Our research focuses on learning reward functions that are sensitive to the numerical values setting. Instead of depending on human feedback or LLM mutation, we suggest an iterative self-alignment that uses LLMs to offer alternative signals that supervise reward learning.

\vspace{-5pt}
\section{Background}

\paragraph{Problem Definition} 

Consider a finite-horizon Markov Decision Process (MDP) parameterized by $(\state, \action, \transition, \reward, \horizon)$ where $\state$, $\action$ are the state and action spaces, $\transition: \state \times \action \to \state$ is the transition function, $\reward: \state \to \mathbb{R}$ is the reward function, and $\horizon$ is the horizon. A policy $\pi$ is a mapping from states to probabilities over actions, $\pi(a|s)$. Given a policy and an MDP, the expected return of the policy is given by $J(\pi;\reward) = \mathbb{E}[\sum_{t=0}^{T-1}\reward(s_t)|\pi]$. The expert policy should be one that optimizes this return, $\pi_E := \arg\max J(\pi; \reward)$ w.r.t. the ground-truth reward $\reward$. In our setting, we are given a partial MDP without i) the reward function $\reward$ nor ii) any forms of expert demonstrations. Instead, we have access to an LLM that can rank a sequence of $M$ trajectories $\traj{k}$ with decreasing preference for $k = 1, \ldots, m$ based on the last state $s^k_T$. The LLM is assumed to have an internal goal or intrinsic understanding of tasks, therefore its ranking is consistent with a human demonstrator optimizing the ground-truth reward function $\reward$.  We aim to find a parameterized reward function $\learnedR$ such that the ranking of these $M$ trajectories based on it is consistent with the ranking given by the LLM.

\vspace{-10pt}
\paragraph{Boltzmann Rationality}{\label{boltzman_model}} 
Similar to modelling noisy optimal human behaviour, we model LLM's preference with a Boltzmann rationality model \cite{luce1959individual}, which assumes it will act to prefer a trajectory with probability proportional to the exponential trajectory return, where $\beta \in [0, \infty)$ is the rationality coefficient controls the level of rationality. Action becomes fully rational and deterministic when $\beta \rightarrow \infty$ and uniformly random when $\beta \rightarrow 0$.

\centerline{$P(\tau) \propto \exp\{\beta\sum R_{\theta}(s_t, a_t)\}$}

Such noisy optimality is critical here as it aligns with our motivation that LLM can be numerically imprecise or unstable. It does not require perfect ranking feedback but only the majority of them. It also aligns with our observation on LLM act rationally but also imperfectly. For instance, it constantly mis-rank in the presence of subtle numerical differences. 
\vspace{-10pt}

\paragraph{Reward Learning from Pairwise Preference} 
In reward learning, denoting $\mathcal{D}$ is the dataset with N pairwise preferences $(\tau_i, \tau_j)$ where $\tau_i \succ \tau_j$, we seek to estimate the true reward parameter $\theta$ that maximizes the posterior: 

\centerline{$P(\theta | \mathcal{D}) \propto \prod_{i=1}^{N}P(\tau_i \succ \tau_j | \theta)P(\theta)$}

Prior is system-dependent and a common choice without special assumption is a uniform prior within the domain $\mathcal{U}[\theta_{min}, \theta_{max}]$. Pairwise preference likelihood $P(\tau_i \succ \tau_j | \theta)$ is modeled with the Bradley-Terry model \cite{bradley1952rank} with Boltzmann-rational model being the score function $p$:
\begin{align}
     P(\traj{i} \succ \traj{j} | \theta) &= 
     \frac{p(\traj{i} | \theta)}{p(\traj{i}|\theta)+p(\traj{j}|\theta)} \\
 &=\frac{\exp\beta\sum\learnedR(s_t^i, a_t^i)}{\exp\beta\sum\learnedR(s_t^i, a_t^i) + \exp\beta\sum\learnedR(s_t^j, a_t^j)}
\end{align}
\vspace{-10pt}

The reward function generated by LLM may be in arbitrary form with complex parameter distribution, sampling-based methods such as the Metropolis-Hastings algorithm \cite{sadigh2017active, hoegerman2023reward} or Gaussian process \cite{biyik2023active} can be deployed to model the posterior flexibly:
\begin{align}
  \label{eqn:argmax_map}
  &\theta_{MAP}(\mathcal{D}) =  {arg\max}_{\theta}\prod_{n=1}^{N}P(\tau_{ni} \succ \tau_{nj} | \theta)P(\theta)\\
  &={arg\max}_{\theta} \sum_{n=1}^{N}\left(\beta\learnedR(\tau_{ni})-log\sum_{\tau \in \mathcal{D}_n}\exp\beta\learnedR(\tau)\right) + \log P(\theta)
  \label{eq:argmin_map}
\end{align}
\vspace{-20pt}

\section{Main Method}

We propose a framework to learn the reward functions from Large Language Models (LLMs). We observe that LLM not only encodes useful task information, it can also serve as a discriminator to evaluate the performance given the observation signals. Based on these, our reward learning method consists of two parts: first we extract the skill-specific reward function parameterization, $\learnedR(.)$, from LLM using a sequence of guiding prompts, then we design an iterative self-alignment procedure to fit the reward function $\learnedR(.)$ using ranking-based preference learning.
\vspace{-5pt}

\subsection{Reward Parameterization Proposal}

We connects LLM to the environment via a dictionary of observations that the environment provides and actions that the agent applies as required by reward calculation. Ideally, a versatile robotics system connects to tasks via modularised interfaces such as a robot interface which define the capabilities of the system. For example inverse kinematics, retrieving object pose, collision detection. They can later utilized to compose tasks and their evaluation. When such structured functions exist, we also feed LLM with an abstraction containing only function define line, type hints with function description, and (input, output) description. In doing so, LLM is expected to leverage upon the existing functions the system is equipped with for reward generation, meanwhile avoiding diluting useful information with a large trunk of function content text or being exposed to any privileged information they may contain.

To generate the initial reward parametrization proposal, we propose to apply the Chain-of-Thought (CoT) to guide the Large Language Model (LLM) in the generation. The details of the prompts we used are included in the appendix. Our process involves a 4-step procedure, 

First, we feed the environment information to LLM with available observations and action descriptions. When interfaces are well defined, we feed the abstract of system interfaces to LLM and ask it to summarize the available functions into a dictionary with the key being the function name, the value being a dictionary summarizing "brief description", "input", and "output".

Next, the task is given as a phrase or short sentence. For example: "touching", or "open the scissor lay flat on the surface with two hands". We prompt the LLM to describe the target skill in natural language. 

In the third step,  LLM is prompted to break the task into steps if it is a multi-step task. For each step, generate a list of ``Dos'' and ``Don'ts'' and identify the relevant observations associated with it. 

Finally, we prompt the LLM to develop the Python function based on analysis from the previous step using observations and functions summarized from the first step. LLM is instructed to put all numerical values into input arguments, and always assign the reward or penalty associated features into a variable before reward calculation. Later the features will be automatically parsed during reward calculation for execution description generation and feedback to LLM.
\vspace{-5pt}

\subsection{Updating Reward via Self-Alignment}



\begin{algorithm}
\caption{Self-Alignment Reward Update}\label{alg:self-alignment}
\begin{algorithmic}[1]

\REQUIRE Learned Reward $\learnedR$, replay buffer $\replay = \varnothing$, initial policy $\pi_{\theta}$
\FOR {$t = 0, 1, ... $} 
    \FOR {$k = 0, 1, ... $} 
        \STATE Update $\pi_{\theta}$ using RL with $\learnedR$ 
    \ENDFOR
    \STATE Sample $M$ trajs $\{\tau_i\}_{i=1,...,M}$using $\pi_{\theta}^*$ 
    \STATE \textit{[optional]} Sample $N$ trajs $\{\tau_i\}_{i=1,...,N}$ evenly from $\replay$ based on reward histogram
    \STATE Calculate \rank{\learnedR} by ranking $\{\tau_i\}$ using \learnedR
    \STATE Calculate \rank{LLM} by ranking $\{\tau_i\}$ using LLM
    \STATE $\mathcal{D}_{neg} \gets$ discrepancy(\rank{\learnedR}, \rank{LLM})
    \STATE $\mathcal{D}_{pos} \gets $ $|\mathcal{D}_{neg}|$ pairs sampled from agreed(\rank{\learnedR}, \rank{LLM})
    \STATE \dataset $\gets \mathcal{D}_{neg}$ + $\mathcal{D}_{pos}$
    \STATE Bayesian update on $\theta$ according to Eq (\ref{eq:argmin_map})
\ENDFOR
\label{algo1}
\end{algorithmic}
\end{algorithm}

The reward is learnt by iteratively optimizing a policy given a reward function in the inner loop, and a reward function in the outer loop with feedback on the policy behaviour. The process iterates and terminates at a satisfactory success rate or total updating steps whichever it reaches earlier. For feedback, ranking is opted instead of the absolute scores or exact numerical adjustments from LLM to suit the restricted numerical reasoning capacity of LLM. For policy learning, model-free Reinforcement Learning (RL) is used due to its flexibility and demonstrated capability for intricate skill learning.

The iterative reward update with self-alignment is presented in Algorithm \ref{algo1}. Within each iteration, the policy is first updated using RL with the current reward function (line 2-4). Next, we draw M samples from the updated policy by rolling it out such that the collected trajectories reflect the current policy behaviour (line 5). When a replay buffer is available such as an off-policy RL is used, the reward histogram of the replay buffer will be parsed and additional N samples are drawn uniformly from the bins (line 6). This is to enable the feedback to be more inclusive so the potential misspecification can be better detected. We then aggregate the samples and retrieve the two ranking sets from the current reward function $R_\theta$ and the LLM through the textual feedback (line 7-8). The textual feedback is automatically constructed by concatenating the identified reward feature names and their values from the reward function with Python local variable parsing. More details on this can be found in Appendix A.2. 

To generate the dataset of pairwise comparison $\dataset = \{ (\traj{a}^0, \traj{b}^0)_0, (\traj{a}^1, \traj{b}^1)_1, \ldots\}$ where $\traj{a}^i \succ \traj{b}^i$, we first parse all inconsistent pairs by comparing the two ranking sets. To resolve the reward inconsistency but also maintain the achieved consistency, we additionally sample an equal amount of consistent pairs from the comparison (line 9 - 10). The ranking from LLM serves as the ground truth and determines $\traj{a}^i \succ \traj{b}^i$ in  $\dataset$. $\dataset$ is further shuffled and further updates the reward model using Bayesian inference. We follow previous works and use Metropolis-Hastings algorithm. The parameters are updated to maximize the posterior of the preference modelling based on the learned reward function with Eq (\ref{eq:argmin_map}). The updated parameter will be accepted if and only if such inconsistency reduces and converges. 

When there is no discrepancy, yet no effective policy is trained after an iteration with a success rate lower than a threshold, the relevant reward or penalty term is assumed to be not salient, LLM is prompted to reflect on the execution and identify the relevant reward or penalty term(s) with corresponding parameter(s). LLM is instructed to suggest a new parameter value where only the value changing direction as + / - is parsed. Such change temporally overwrites the parameter domain from $[\theta_{\text{min}}, \theta_{\text{max}}]$ to $(\theta_{\text{current}}, \theta_{\text{max}}]$ or $[\theta_{\text{min}}, \theta_{\text{current}})$ and conduct new parameter search with self-alignment. Similarly, the updated parameter will be accepted with maximumly reduced and no new inconsistency generated. In doing so, we combine heuristics as active reward adjustment while maintaining the reward logic by preserving the reward rank to avoid unintentional flaws being introduced with the new parameterization. 
\vspace{-5pt}

\section{Experiments}

\subsection{Evaluation Objectives}

To evaluate the proposed framework, we design our experiments to answer the following three questions:

\textbf{\textit{$\mathcal{H}1$ -}} Can this pipeline generate effective reward functions to induce optimal policies on varied skills learning?

\textbf{\textit{$\mathcal{H}2$ -}}  Can the periodic update through self-alignment improve the numerical impreciseness and instability, thus the efficacy of reward functions?

\textbf{\textit{$\mathcal{H}3$ -}}  How does this method perform compared to the alternative unsupervised update method, i.e. through LLM reflected mutation and rejection? 

\vspace{-5pt}
\subsection{Evaluation Tasks}

\begin{figure}[h]
\centering
\begin{minipage}[b]{.333\linewidth}
\includegraphics[width=\textwidth]{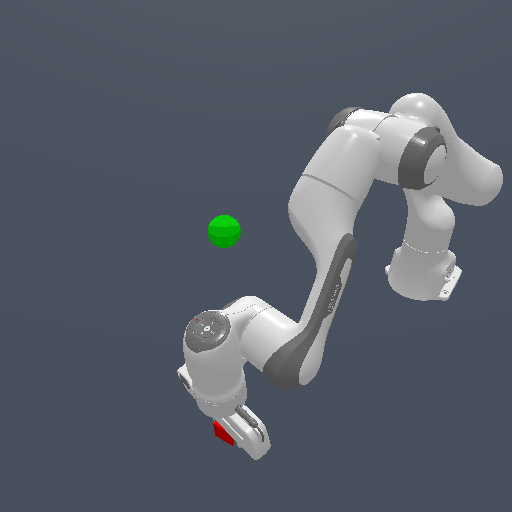}
\label{fig:pick_cube_env}
\end{minipage}\hfill
\begin{minipage}[b]{.333\linewidth}
\includegraphics[width=\textwidth]{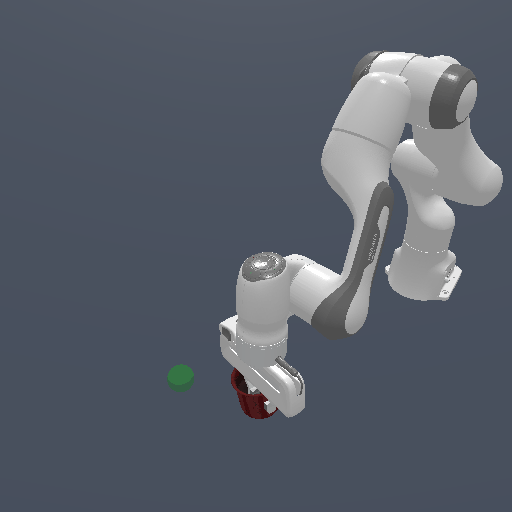}
\label{fig:pick_ycb_env}
\end{minipage}\hfill
\begin{minipage}[b]{.333\linewidth}
\includegraphics[width=\textwidth]{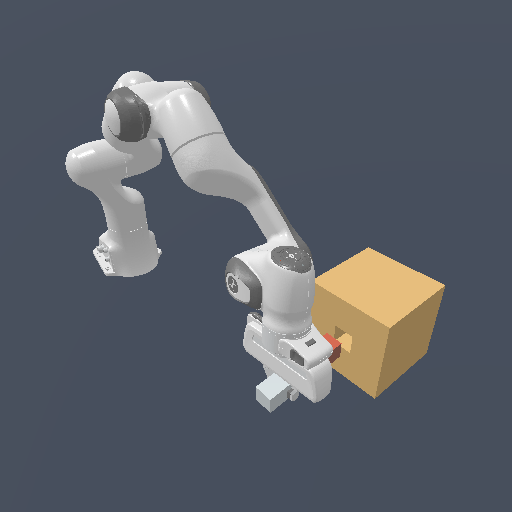}
\label{fig:peg_insertion_env}
\end{minipage}\hfill
\vspace{-13pt}
\begin{minipage}[b]{.333\linewidth}
\includegraphics[width=\textwidth]{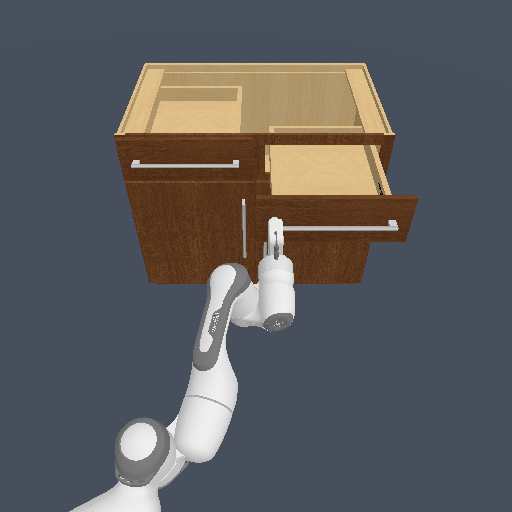}
\label{fig:open_drawer_env}
\end{minipage}\hfill
\begin{minipage}[b]{.333\linewidth}
\includegraphics[width=\textwidth]{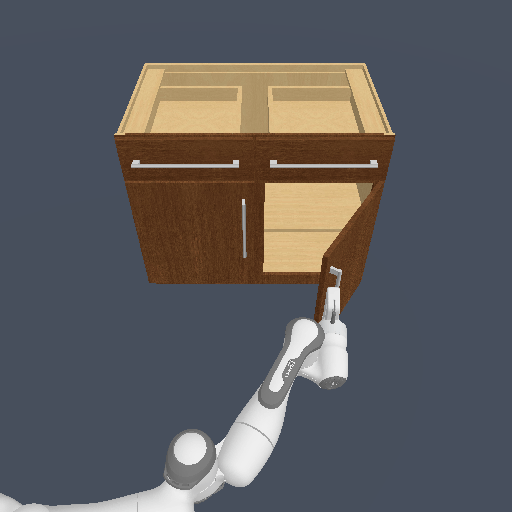}
\label{fig:open_door_env}
\end{minipage}\hfill
\begin{minipage}[b]{.333\linewidth}
\includegraphics[width=\textwidth]{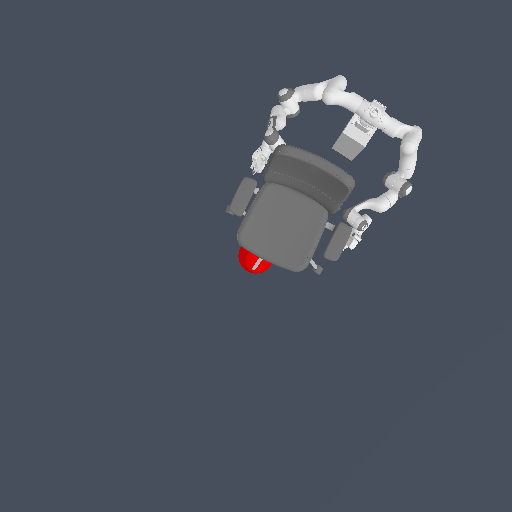}
\label{fig:push_chair_env}
\end{minipage}\hfill
\vspace{-28pt}
\caption{Six evaluation tasks from ManiSkill2: PickCube, PickSingleYCB, PegInsertionSide, OpenCabinetDoor, OpenCabinetDrawer, PushChair.}
\label{fig:maniskill2_evaluation_tasks}
\end{figure}

\begin{figure}[h]
\centering
\begin{minipage}[b]{.33\linewidth}
\includegraphics[width=\textwidth]{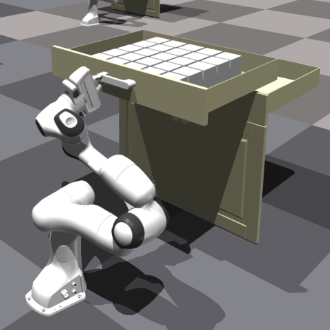}
\label{franka_cabinete_env}
\end{minipage}\hfill
\begin{minipage}[b]{.33\linewidth}
\includegraphics[width=\textwidth]{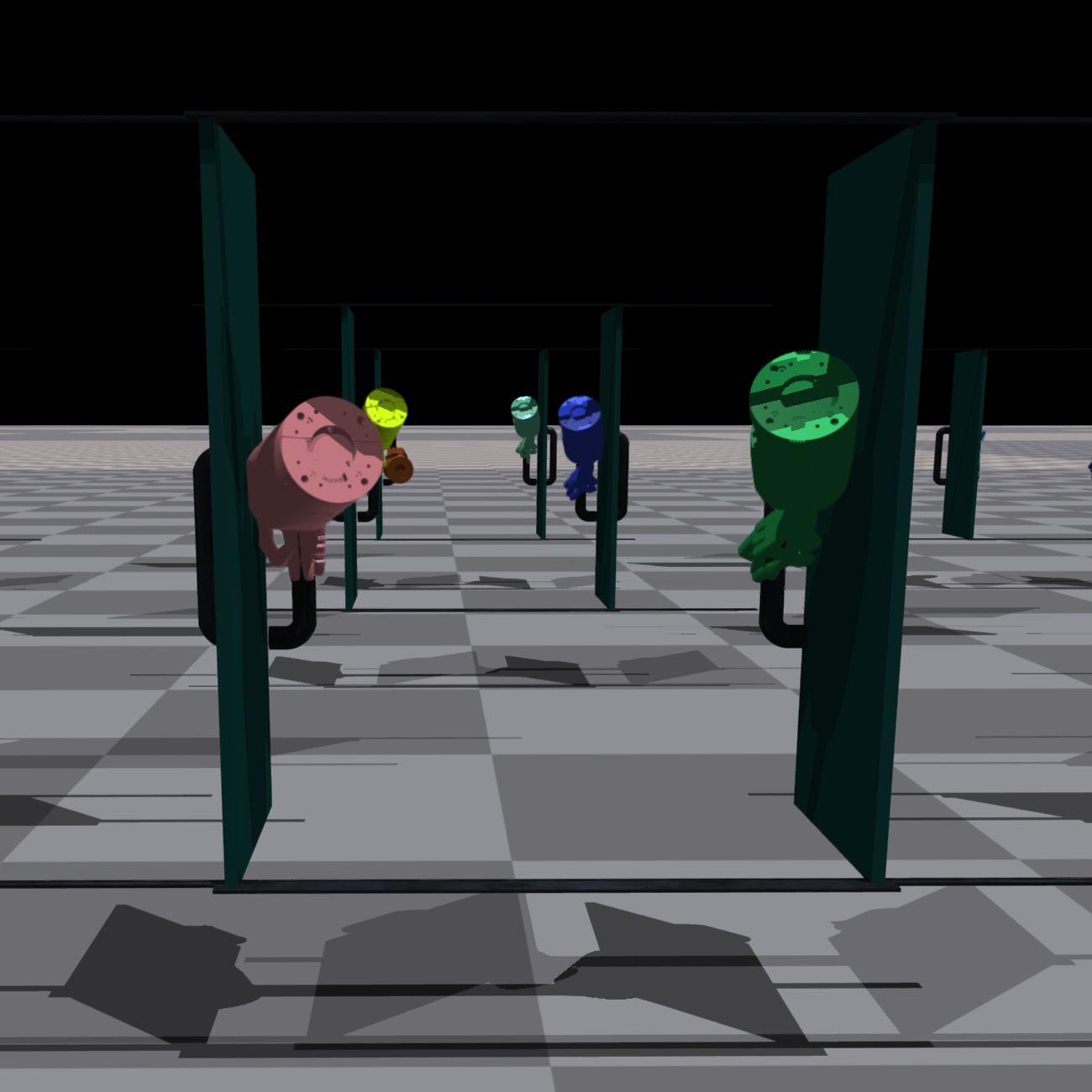}
\label{push_door_outward_env}
\end{minipage}\hfill
\begin{minipage}[b]{.33\linewidth}
\includegraphics[width=\textwidth]{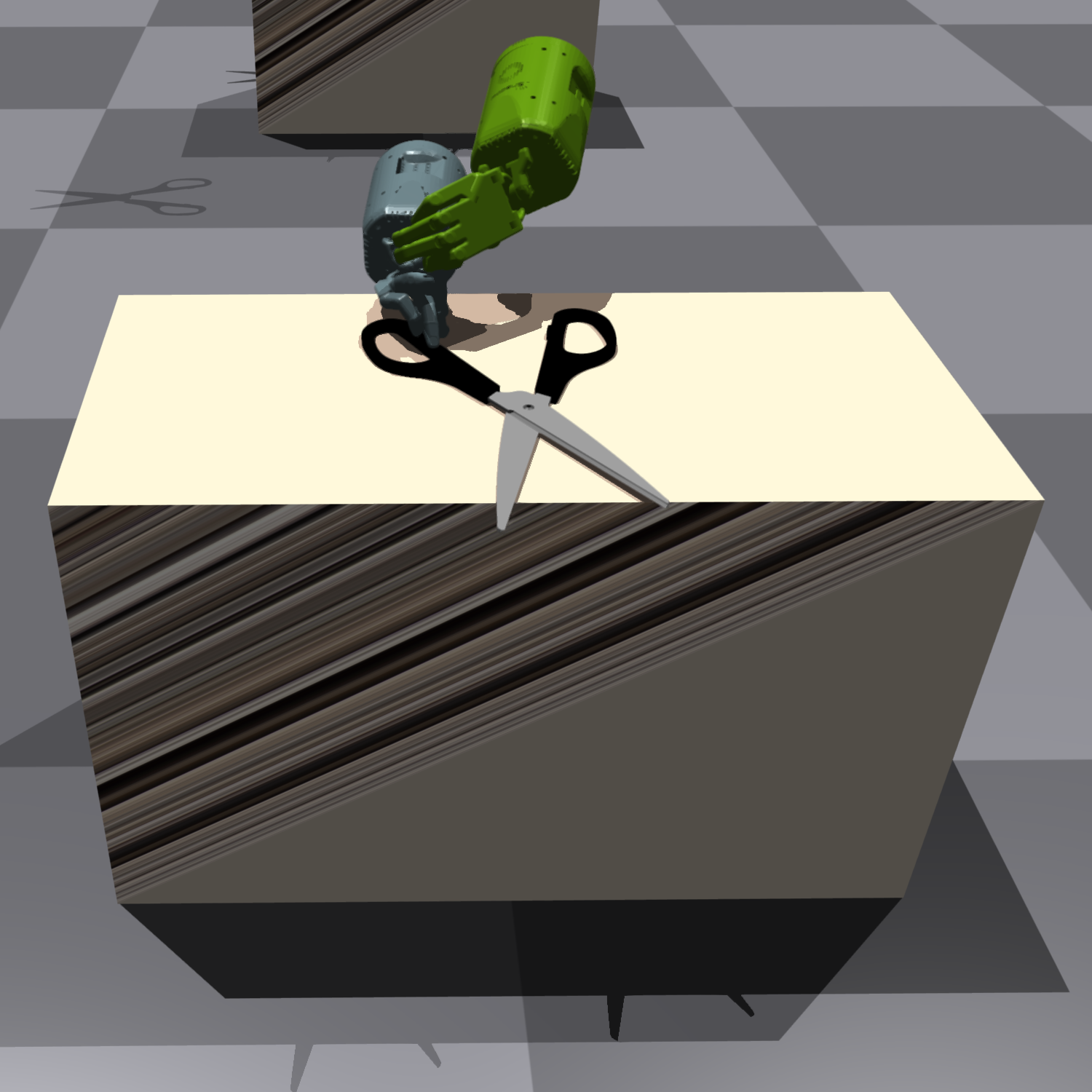}
\label{open_scissor_env}
\end{minipage}\hfill
\vspace{-15pt}
\caption{Three Isaac Gym evaluation tasks: Franka Cabinet, Shadow Hand Open Door Outwards, Shadow Hand Open Scissor}
\label{fig:isaac_evaluation_tasks}
\end{figure}
\vspace{-5pt}

Following the objective, we evaluated our framework on 6 manipulation tasks in ManiSkill2 \cite{gu2023maniskill2} as illustrated in Fig \ref{fig:maniskill2_evaluation_tasks}. The tasks includes rigid and articulated object manipulation with fixed-based manipulator, single-arm and dual-arm mobile manipulator. 

\subsection{Baselines}

We compare our learnt reward function with two other reward functions. One trains policy with expert-designed oracle rewards from the original environment implementations, through which we examine the efficacy of our learnt reward function for $\mathcal{H}1$. Another trains policy with the intially proposed reward function from LLM analysed with CoT, but the parameterization stays fixed throughout the training. Through this comparison we aim to examine the efficacy of self-alignment update for $\mathcal{H}2$.

We also compare our method with two baseline approaches of reward generation with LLM: \textit{Text2Reward} \cite{xie2023text2reward} on four overlapping ManiSkill2 tasks and \textit{Eureka} \cite{ma2023eureka} on three tasks implemented with Isaac Gym \cite {makoviychuk2021isaac}.

\begin{itemize}
    \vspace{-10pt}
    \item \textbf{\textit{Text2Reward}} generates dense reward generation with LLM by similarly providing the Pythonic environment abstraction and task description to LLM. The reward function can be further updated iteratively with LLM by taking human-feedback.
    \vspace{-5pt}
    \item \textbf{\textit{Eureka}} generates high-performance reward function in an evolutionary manner. It feeds the environment script to LLM and generates multiple reward functions per time for policies training in-parallel. The batch success rates are fed back to LLM for reward functions' reflection and mutation then trains the next batch policies. Such loop is iterated until meets the termination condition. 
    \vspace{-10pt}
\end{itemize}

We performed the self-alignment update on reward functions generated without human feedback in \textit{Text2Reward} to examine if the improvement of such self-alignment update for $\mathcal{H}2$ holds, does it still hold for reward functions generated externally. We also compare the training and token efficiency with alternative unsupervised method proposed in \textit{Eureka} for $\mathcal{H}3$ on 3 tasks shown in Fig \ref{fig:isaac_evaluation_tasks}. 

\vspace{-5pt}
\subsection{Training Setup}

For policy training, we adopted the same RL implementation and hyper-parameter setting that the baseline methods used, correspondingly are SAC \cite{haarnoja2018soft} from stable-baselines3 \cite{stable-baselines3} and PPO \cite{schulman2017proximal} from rl\_games \cite{rl-games2021}. The success rate threshold to query parameter adjustment is 50\% for SAC and 10\% for PPO. For GPT, we used GPT-4 with API model name gpt-4-0613. As gpt-4-0314 used in \textit{Eureka} is deprecated, we validated over 10 iterations that the new model is able to generate $11.00\pm1.34$ out of 16 successfully executed reward functions compared to $9 / 16$ from the author thus should not degenerate its performance. 

For reward update, we set the rationality coefficient $\beta = 0.9$, feedback at every $10000$ training steps for ManiSkill2, 100 epochs for Isaac Gym with M=5 roll-out samples from the latest policy and N=5 for sampling from the reward histogram. The training step equals the exploration step in our setup. For reward update with Metropolis-Hastings algorithm, our customized implementation is built upon APReL \cite{biyik2022aprel}. The burn-in period is 200 iterations and the number of samples is 100. The proposal distribution follows Gaussian distribution as $\mathcal{N}(\bar{\theta}, 0.2)$ on normalized parameters then is clipped to [0, 1]. 

In practice, as the reward parameters may be multi-modal \cite {biyik2023active} thus multiple sets of values may fulfil the same level of ranking discrepancy reduction. To minimize the reward fluctuation over iterations, we model the posterior and update the reward by sampling in the feasible region within the distances to the current value of [1.0, 3.0, 5.0, 10.0] in parallel. This setting works for parameter value ranges from $\pm e^{-2} - e^1$ scale proposed by LLM in our experiments thus no additional tuning was implemented and may be further explored. The new parameters with a maximum discrepancy reduction will be accepted. If the same level of discrepancy reduction is achieved, we select the set of values that are the nearest to the current one.

\vspace{-5pt}
\subsection{Results}

For evaluation, we report the success rates over five different seeds against the exploration steps on 6 ManiSkill2 manipulation tasks. The results are plotted Fig \ref{fig:maniskill2}. We also conducted similar reward learning on reward functions that are generated from \textit{Text2Reward} zero-shot cases with LLM. The comparison on policy trained with fixed-parameterized reward function and updated with self-alignment are plotted in Fig \ref{fig:text2reward_comparison}.

\subsubsection{The learnt reward is able to induce optimal policy} 

We first analyse through objective $\mathcal{H}1$. Compared to the oracle reward, we observed that our method is able to consistently develop a policy close to policy trained with expert-designed oracle reward  across all 6 ManiSkill2 tasks. In the case with a better reward design proposed by LLM in PushChair task, we are able to further push the performance from $60.59\%$ to $83.65\%$ with oracle reward gives at the peak $35.34\%$.

\subsubsection{Self-Alignment Update Consistently Improves Training Efficacy and Efficiency} 
\begin{figure*}[h]
\centering
\begin{subfigure}
  \centering
  \includegraphics[width=0.31\linewidth]{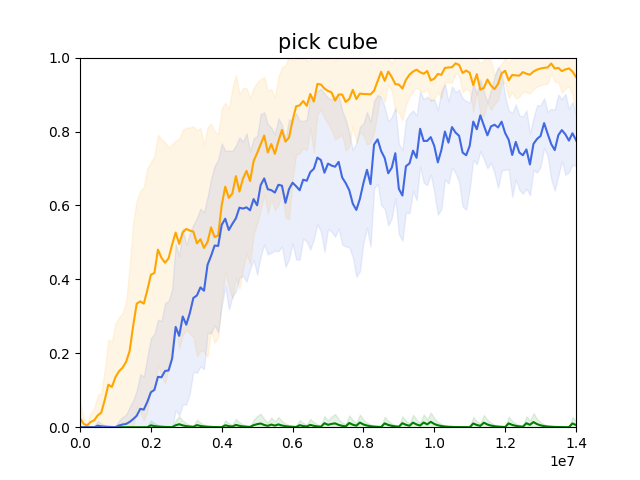}
  \label{fig:pick_cube}
\end{subfigure}
\begin{subfigure}
  \centering
  \includegraphics[width=0.31\linewidth]{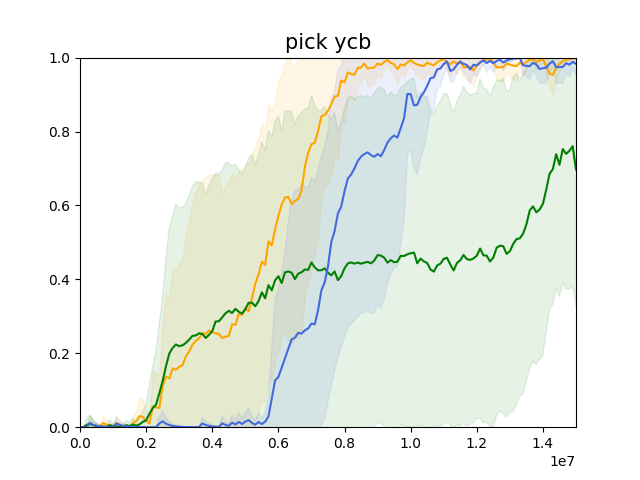}
  \label{fig:pick_ycb}
\end{subfigure}
\begin{subfigure}
  \centering
  \includegraphics[width=0.31\linewidth]{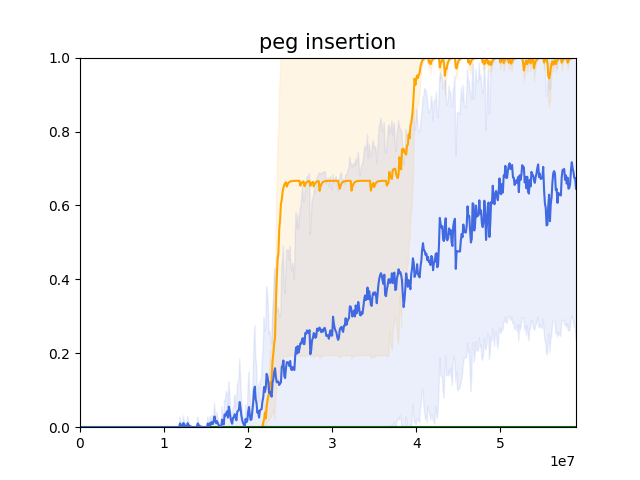}
  \label{fig:peg insertion}
\end{subfigure}
\begin{subfigure}
  \centering
  \includegraphics[width=0.31\linewidth]{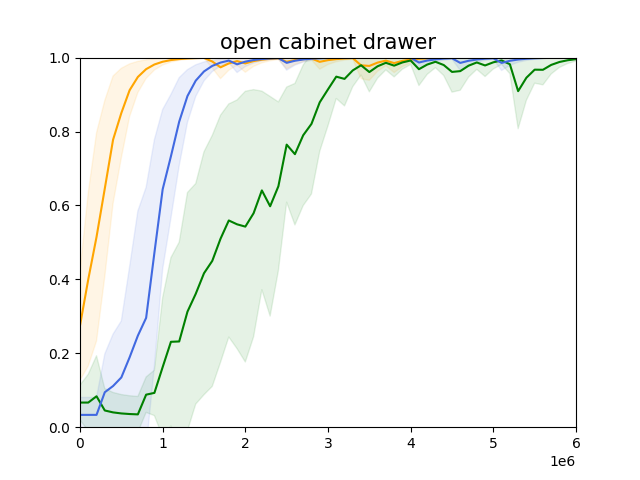}
  \label{fig:open_drawer}
\end{subfigure}
\begin{subfigure}
  \centering
  \includegraphics[width=0.31\linewidth]{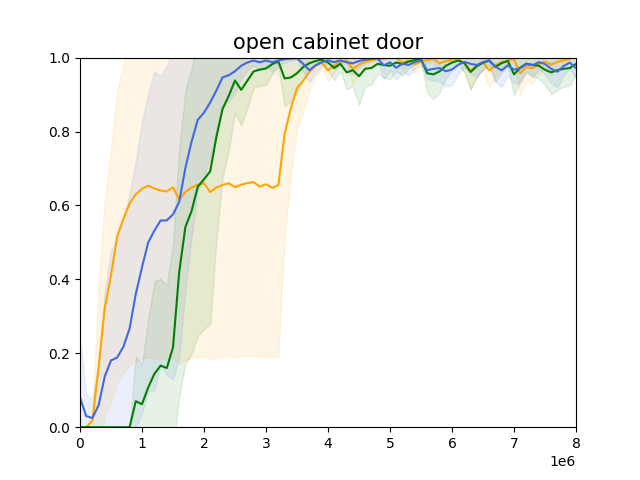}
  \label{fig:open_door}
\end{subfigure}
\begin{subfigure}
  \centering
  \includegraphics[width=0.31\linewidth]{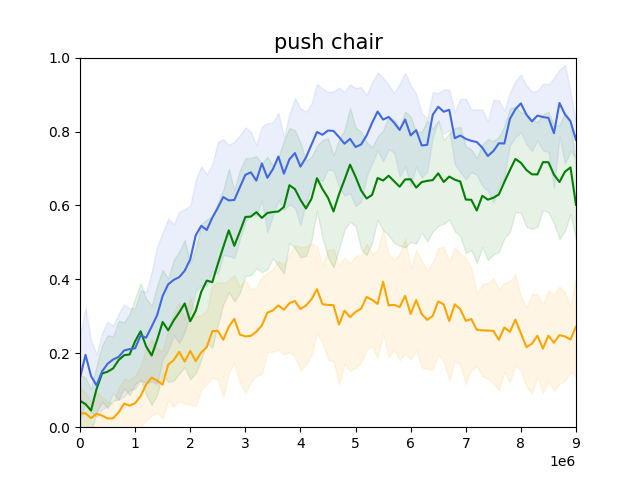}
  \label{fig:push_chair}
\end{subfigure}
\includegraphics[width=0.5\linewidth]{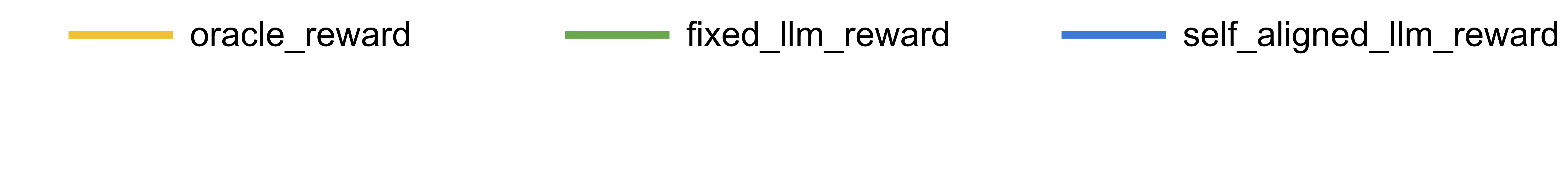}
\vspace{-25pt}
\caption{Success rates vs exploration steps on 6 ManiSkill Tasks with SAC. The updated reward is able to produce policy with similar performance to that is trained with oracle reward on 5 tasks. Compared to using fixed reward function genreated by LLM, our approach consistently improves the training with faster convergence rate and/or higher convergence performance }
\vspace{-5pt}
\label{fig:maniskill2}
\end{figure*}

\begin{figure*}[h]
\centering
\begin{subfigure}
  \centering
  \includegraphics[width=0.23\linewidth]{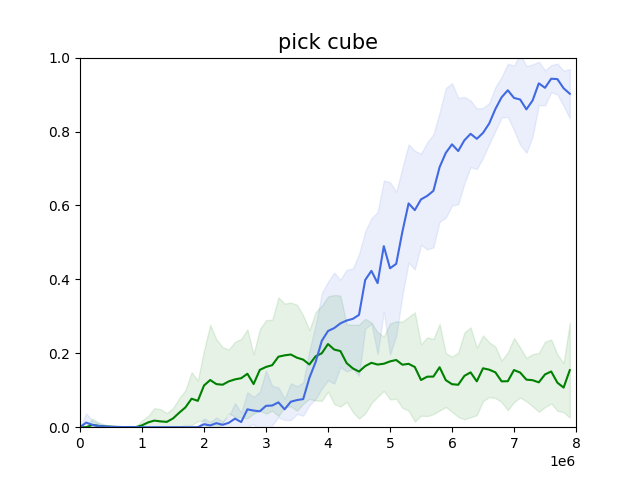}
  \label{fig:t2r_pick_cube}
\end{subfigure}
\begin{subfigure}
  \centering
  \includegraphics[width=0.23\linewidth]{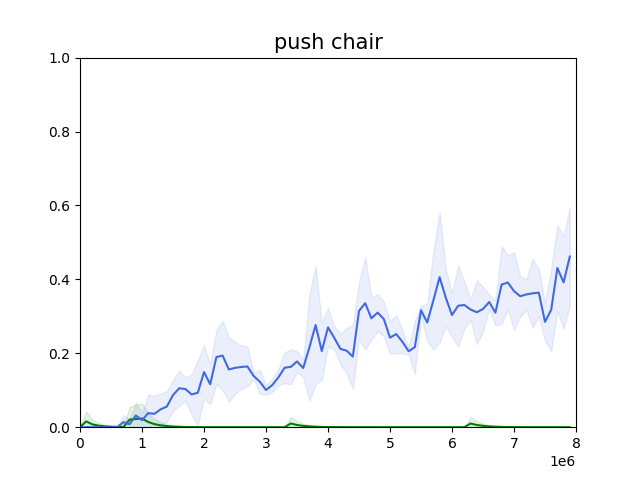}
  \label{fig:t2r_push_chair}
\end{subfigure}
\begin{subfigure}
  \centering
  \includegraphics[width=0.23\linewidth]{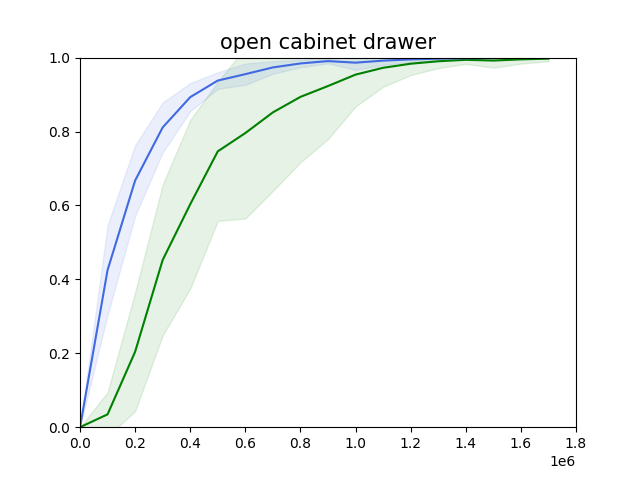}
  \label{fig:t2r_open_drawer}
\end{subfigure}
\begin{subfigure}
  \centering
  \includegraphics[width=0.23\linewidth]{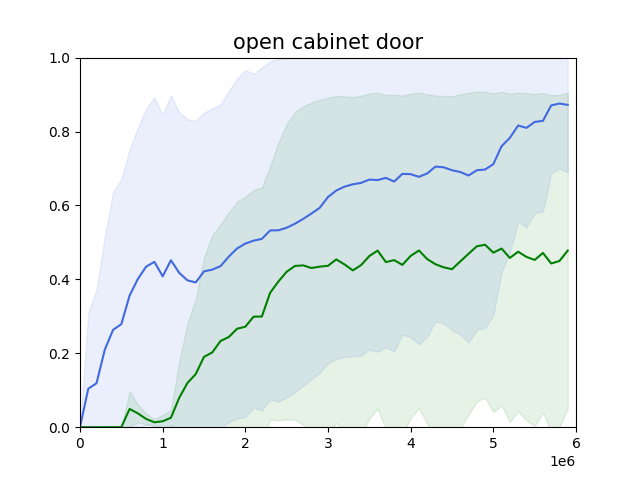}
  \label{fig:t2r_open_door}
\end{subfigure}
\vspace{-6pt}
\includegraphics[width=0.5\linewidth]{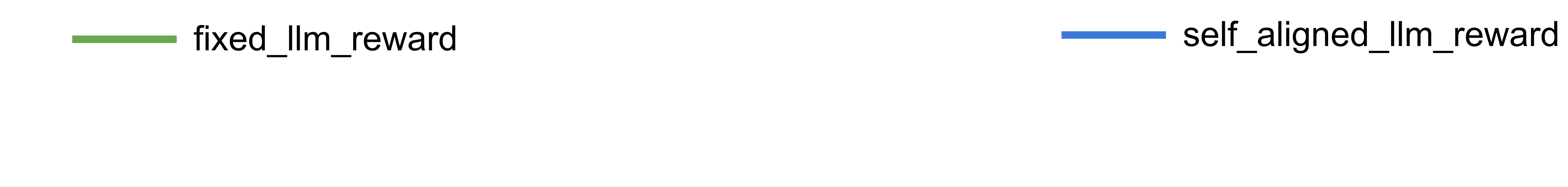}
\vspace{-20pt}
\caption{Success rate vs exploration steps with Text2reward zero-shot reward functions with SAC. Similarly, the periodically update reward function through self-alignment consistently improves the final performance achieved.}
\vspace{-15pt}
\label{fig:text2reward_comparison}
\end{figure*}

From Fig \ref{fig:maniskill2}, it can be observed that LLM generated reward function can often lack of adequate numerical optimality to develop the optimal policy as indicated in our motivation and $\mathcal{H}2$. This can be demonstrated by the nearly zero success rate of policies trained with fixed-parameterized reward proposed by LLM for tasks Pick-Cube and Peg-Insertion. There are the same patterns for Pick-Cube and Push-Chair tasks on functions from \textit{Text2Reward} in Fig \ref{fig:text2reward_comparison} which is consistent with the original paper. 

For all cases, the self-alignment update scheme effectively improves the performance reflected as (1) faster convergence; (2) higher success rate at convergence in general. We think the such effectiveness or improvement are mainly due to two reasons:

\textit{(1) LLM brings meaningful inductive bias to steer policy towards global-goal especially in multi-objective setting}:

\begin{figure}[ht]
\centering
\begin{minipage}[b]{.5\linewidth}
\includegraphics[width=\textwidth]{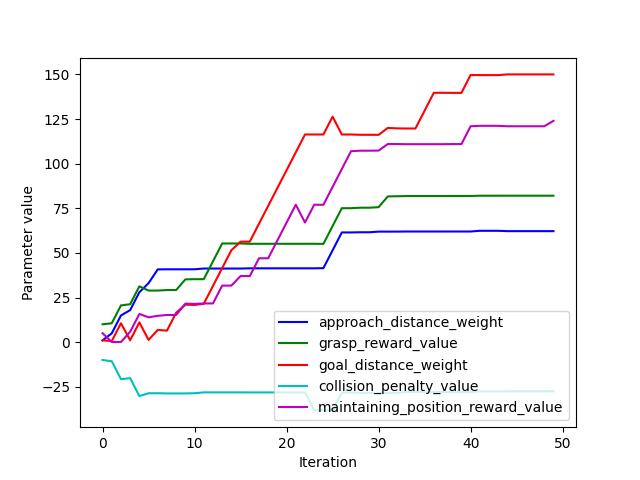}
\end{minipage}\hfil
\begin{minipage}[b]{.5\linewidth}
\includegraphics[width=\textwidth]{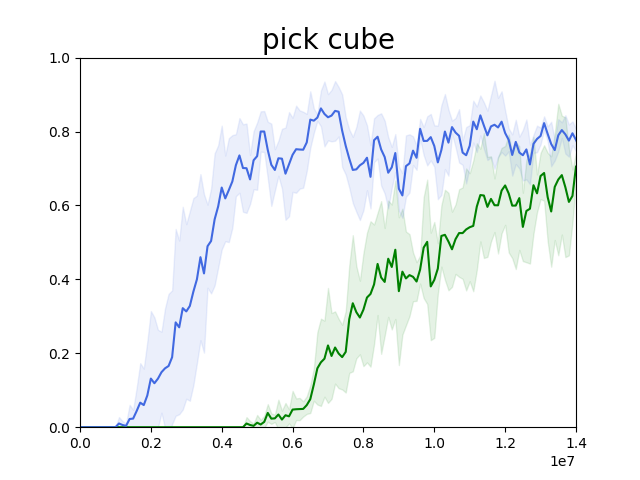}
\end{minipage}
\vspace{-15pt}
\caption{(a) parameter update over iteration for pick cube task; To better visualize the early shift the update is truncated to 50 iterations. (b) success rate of policy trained with the \text{\color{blue}actively adjusted} reward function and with the \text{\color{teal}adjusted final} reward function. The policy may not achieve the same performance trained with the final reward function learnt only.}
\vspace{-5pt}
\label{weight_curriculum}
\end{figure}

Following the prompt, LLM decomposes task objectives as Do's and Don't step by step into dense reward and penalty terms. Such multi-objective learning especially for multi-step tasks can be challenging to learn altogether. By feedback on recent exploratory behaviors from the replay and exploitative behavior from current policy to LLM, LLM identifies the most and least optimal behaviors under the local context and steers it towards the global goal. 

Such behavior is similar to introducing a curriculum for learning with periodic feedback and can be observed from the weight parameter change. For instance, for PickCube or PegInsertion, reward update were emphasized on encouraging to approach first, later when contact starts with policy improving with \texttt{grasped=True} from the feedback samples start to appear, the weights further shift to increase grasping weights to encourage stable grasping. It is followed to increase transporting to goal pose weight with varied \texttt{distance\_to\_goal} in feedback samples. The pattern can be observed in Fig \ref{weight_curriculum} and Fig \ref{fig:transition_visualization} in Appendix. By iteratively doing so, the final policy was effectively guided toward the final target. We think this might be the main reason for this method to yield a faster convergence rate.

\textit{(2) Execution feedback and self-alignment prevents sub-optimal behavior due to parameter mis-specification:}

As indicated in numerous existing literature, reward shaping is nontrivial as inappropriate shaping will introduce unintentional rewards or penalties that further leads to sub-optimal policy by exploiting such flaw. Meanwhile as we observed, the way LLM generates the reward code with parameterization directly as text generation task is not the best way as reward shaping requires. Through a closed-loop reward update via self-alignment, the shaping logic is enforced through the ranking further as pairwise preference. Such numerical errors or flaws can be reflected through execution feedback and ranking discrepancy that are later corrected via posterior maximization. 

\begin{figure*}[h]
\centering
\begin{subfigure}
  \centering
  \includegraphics[width=0.31\linewidth]{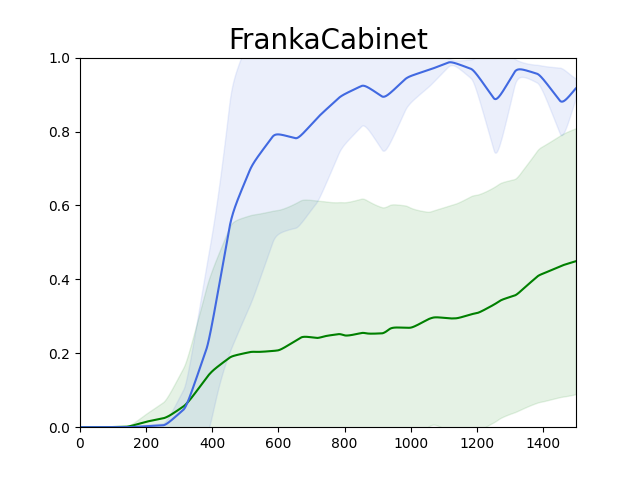}
  \label{fig:isaacgym_franka_cabinet}
\end{subfigure}
\begin{subfigure}
  \centering
  \includegraphics[width=0.31\linewidth]{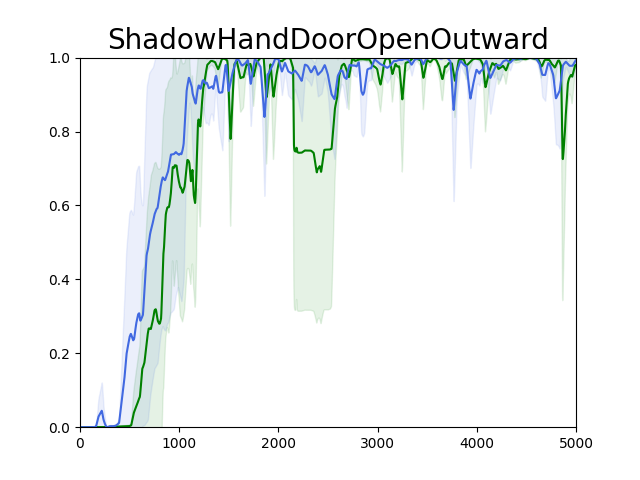}
  \label{fig:isaacgym_shadow_door}
\end{subfigure}
\begin{subfigure}
  \centering
  \includegraphics[width=0.31\linewidth]{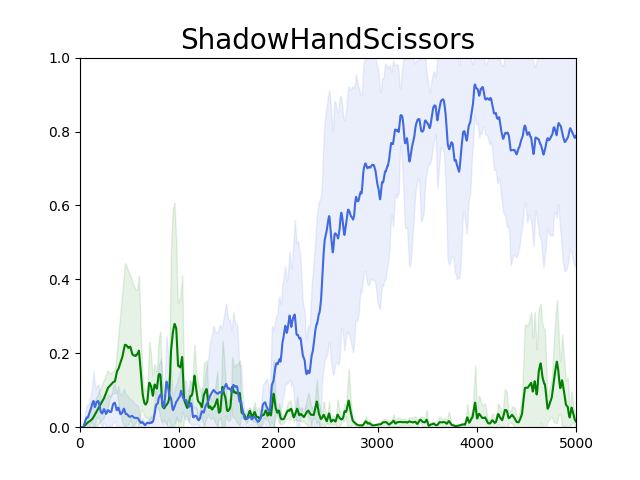}
  \label{fig:isaacgym_shadow_scissor}
\end{subfigure}
\vspace{-5pt}
\includegraphics[width=0.5\linewidth]{figs/maniskill/legend_2.png}
\vspace{-20pt}
\caption{Success rate with PPO policy on 3 IsaacGyms tasks: (a) Franka Cabinet; (b) Push Door Outwards; (c) Open Scissor. }
\vspace{-15pt}
\label{isaacgym_exp}
\end{figure*}

For example, in the Open-Cabinet-Drawer/Door task, when given a high reach reward and relatively lower pull reward, the policy converges to stay in touch with the handle, as the poor pulling action easily loses contact with the handle and yields a lower reward. However, LLM ranks with a preference for a larger pulling distance for later-stage action. The reward is updated to increase the pulling weight iteratively to correct such sub-optimality and move towards more pushing actions. 

\begin{figure}[ht]
\centering
    \includegraphics[width=0.48\textwidth]{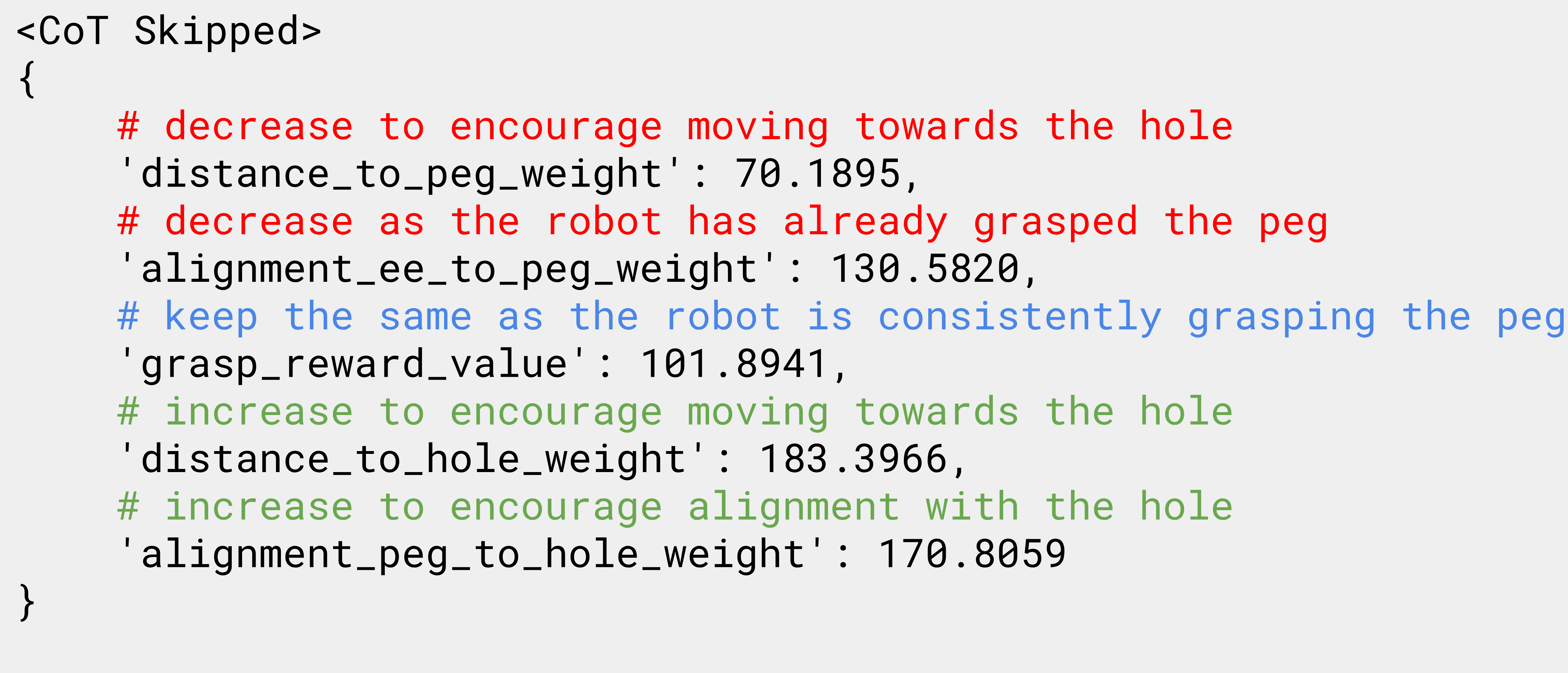}
\vspace{-15pt}
\caption{Parameter adjustment for Peg-Insertion task from LLM with execution feedback. LLM proposed to reduce approach weight to better encourage the final goal reaching of the peg.}
\label{reduce_to_explore}
\end{figure}

It is also a common pattern across tasks that LLM will recommend to reduce weight for a behaviour that is well-established for better exploration to develop the next step behaviour. One example taken from the Peg Insertion task is shown in Fig \ref{reduce_to_explore}.

\vspace{-5pt}

\subsubsection{Significant Lower Token Consumption Compared to Mutation-Based Method} 

Lastly, we report the comparison between our method with \textit{Eureka}. The success rates for training with fixed and updated reward functions are plotted in Fig \ref{isaacgym_exp}. We also compare the \texttt{n\_tokens} as the total input and output tokens consumed \cite{hu2023tree} at termination in table \ref{gpt_token}, which is either the training reaches the 1.0 success rate or 50 feedback iterations to prevent uncapped high cost for querying GPT-4.  

\begin{table}[ht]
    \begin{center}
    \begin{tabularx}{0.48\textwidth}{
    >{\centering\arraybackslash}X 
    |>{\centering\arraybackslash}X 
    >{\centering\arraybackslash}X 
    >{\centering\arraybackslash}X }
     \hline
     \textit{n$\_$tokens}$\downarrow$ & \textit{Cabinet} & \textit{Door} & \textit{Scissor} \\ 
     \hline
     \textit{Eureka} & 64,698\quad(\textbf{3}) & 1,284,924 (-) & 1,338,778 (-)\\ 
     \textbf{Ours} & \textbf{15,633} (7) & \textbf{17,543} (\textbf{7}) & \textbf{130,969(-)} \\
     \hline
    \end{tabularx}
    \vspace{-10pt}
    \caption{Total number of tokens consumed throughout training. (n) indicates the number of iterations where success rate = 1.0 is achieved. (-) means termination at max iteration 50.}
    \label{gpt_token}
    \end{center}
\end{table}
\vspace{-15pt}
\begin{table}[ht]
    \begin{center}
    \begin{tabularx}{0.48\textwidth}{
    >{\centering\arraybackslash}X 
    |>{\centering\arraybackslash}X 
    >{\centering\arraybackslash}X 
    >{\centering\arraybackslash}X }
     \hline
    \textit{max sr} $\uparrow$ & \textit{Cabinet} & \textit{Door} & \textit{Scissor} \\ 
     \hline
     \textit{Eureka} & \textbf{100.00$\%$} & $0.00\%$  & $0.24\%$\\ 
     \textbf{Ours} &  \textbf{100.00$\%$} & \textbf{100.00$\%$} & \textbf{94.54$\%$} \\
     \hline
    \end{tabularx}
    \vspace{-10pt}
    \caption{Maximum success rate achieved throughout training.}
    \label{max_sr}
    \end{center}
\end{table}

\vspace{-10pt}
On the three tasks, \textit{Eureka} on average consumes tokens about 10 times of our method used per iteration. However, the total tokens consumed throughout training thus the cost can reach up to 100 times. We believe the evolutionary method in \textit{Eureka} is a nice way to address the error-prone nature of code generated from LLM. However, it may not be efficient for parameterization setting. In addition, \textit{Eureka} iterates in a relatively aggressive way which may also be related to the sample efficiency. The policies are trained for 5 epochs before the success rate feedback and retraining, where we observe hundreds to thousands of epochs are often required for visible progress on success rate across three tasks. This might be more pronounced with sparser-reward tasks. Lastly, though LLM mutates through heuristic reflection, it does not ensure such mutated reward will perform better rather than fluctuate which was observed for open door and scissor tasks as shown in table \ref{max_sr}. Through self-alignment, such objective is quantified in the ranking discrepancy that aligns with the objective of reward design, and optimized towards a reduction in such measure. 

\subsection{Ablation Study}
To better understand how individual components, i.e. self-alignment (SA) and active adjustment (AA) of parameters contribute to the final performance, we conducted the ablation study that updates the reward parametrization by: 
\vspace{-5pt}
\begin{enumerate}
    \vspace{-5pt}
    \item enforcing the ranking self-consistency only.
    \vspace{-5pt}
    \item actively adjusting through LLM reflection only.
\end{enumerate}
\vspace{-10pt}
The reward functions are updated iteratively at the same frequency. We analyse the effects on three tasks: Pick-Cube (16M steps), Open-Cabinet-Door (8M steps) and Push-Chair (8M steps) and report the final policy success rate averaged over 5 seeds in table \ref{ablation} below:

\begin{table}[ht]
    \begin{center}
    \begin{tabularx}{0.48\textwidth}{
    >{\centering\arraybackslash}X 
    |>{\centering\arraybackslash}X 
    >{\centering\arraybackslash}X 
    >{\centering\arraybackslash}X 
    >{\centering\arraybackslash}X }
     \hline
    \textit{sr} $\uparrow$ & \textit{Fixed} & \textit{SA Only} & \textit{AA Only} & \textit{SA + AA}\\ 
     \hline
     \textit{Cube} & 0.01$\pm$0.02 & 0.30$\pm$0.15  & 0.47$\pm$0.49 & \textbf{0.73$\pm$0.16}\\ 
     \textit{Cabinet} & 0.97$\pm$0.05 & 0.97$\pm$0.03 & \textbf{0.99$\pm$0.02} & \textbf{0.99$\pm$0.02}\\ 
     \textit{Chair} & 0.69$\pm$0.16 & 0.74$\pm$0.14  & 0.76$\pm$0.14 & \textbf{0.84$\pm$0.10}\\ 
     \hline
    \end{tabularx}
    \vspace{-10pt}
    \caption{Success rate for the final policies trained with (1) fixed LLM reward; (2) updated reward with self-alignment only; (3) updated reward with active parameter adjustment only; (4) updated reward that uses both strategies as proposed in this paper.}
    \label{ablation}
    \end{center}
\end{table}
\vspace{-5pt}

It is observed that both schemes improved the performance compared to using a fixed LLM reward across the three tasks, while combining the two strategies consistently yields the highest improvement. Actively adjusting the parameters in general leads to better policies compared to only enforcing the reward ranking self-consistency. This is aligned with the underlying mechanisms of the two strategies that the self-alignment aims to prevent error from potential reward misspecification, while active parameter adjustment aims to directly improve the reward saliency. However, with the task being more challenging or longer horizon which takes longer to establish the target behaviour, the active parameter adjustment was observed to produce drastic changes and value differences among parameters with performances stopped improving over iterations. 

\vspace{-5pt}
\subsection{Limitations and Future Works}
In this section, we analyse a few limitations in this method and potential future works. First, to summarize the execution, the current method reports using the reward features from the last step, as concatenation over steps will be lengthy for 2/500-step tasks and dilute the useful information. Thus it works for tasks where execution can be evaluated based on the final configuration, but not where trajectory matters such as walking in an S shape. With the development of a more powerful VLM, we hope the feedback will be more informative to remove such limitations. Another limitation lies in the preference ranking. Currently, we assume all rankings are equally better or worse, However, for sparse-reward tasks such as peg insertion, some successful experiences are significantly better and integrating it will enable more efficient training, which we hope to address in future work. 

\vspace{-5pt}
\section{Conclusion}
In this work we designed a framework to utilize LLM to propose the features and parameterization of reward function, and iteratively update the parameters through a self-alignment process in the absence of human intervention. More specifically, we periodically feedback policy to LLM for ranking. The ranking is later served as pairwise preference to align and update the reward function. When the ranking is fully aligned yet no effective policy is developed, we also query LLM for active reward parameter adjustment under the same framework of self-alignment. We validated the framework in 6 ManiSkill2 tasks and 3 IsaacGym tasks and shows the framework is able to induce optimal policy effectively and efficiently, where it can often fail without such self-alignment update. The method is also proven to be significantly token efficient comparing to alternative mutation-based method.

\pagebreak
\section*{Impact Statement}

This paper contributes to the field of robotics and machine learning by introducing a method that enhances the learning of reward functions in robots automatically, utilizing Large Language Models (LLMs). While our research primarily focuses on technical advancements, we acknowledge its broader societal implications. The refinement of reward functions in autonomous systems has the potential to benefit sectors like healthcare, manufacturing, and services by improving efficiency and reliability.

We recognize the ethical importance of ensuring that autonomous systems operate in alignment with societal norms and values. Our work aims to improve the precision of reward functions and better alignment of social preference that could potentially regulate behaviors including unethical or socially biased behaviors. It can produce a more predictable and trustworthy robotic behavior. We emphasize the need for ongoing interdisciplinary dialogue to address the ethical challenges and ensure that advancements in robotics and machine learning are developed and applied responsibly, aligning with broader societal interests.

\nocite{langley00}

\bibliography{example_paper}
\bibliographystyle{icml2024}

\newpage
\appendix
\onecolumn
\section{Appendix}
\subsection{Prompt Details}
\subsubsection{Background Prompt}
First, we list the prompt on the task training background using reinforcement learning and available observations as the fixed observation dictionary that LLM can access. An additional wrapper was added for ManiSkill2 tasks to create such a fixed observation dictionary as such information is scattered in the original environment script. By doing this, LLM can be isolated from the environment source code and communicate with it through only observation and action. This is to mimic a fixed robotic system equipped with adequate sensory information that is expected to be versatile and acquire different skills when given the right rewards:

\begin{lstlisting}[breakautoindent=true, breakindent=10pt, breaklines]
Our goal is to learn a set of sensory-motor skills from language concepts with multi-modal sensory observations with robots interacting with the environment and learn through trial and error in a reinforcement learning manner.  In the environment, we have:
(1) a robot (access by env.robot) which we can communicate and control using the robot interface as showed previously;
(2) a target object (env.target_object) to be manipulated. None if the task does not involve target object.

The available observations from the environment are stored in a dictionary obs which includes:
1. obs['joint_positions']: robot joint positions;
2. obs['joint_velocities']: robot joint velocities;
3. obs['ee_pose']: end effector pose;
4. obs['ee_contact_force']: the external contact force at the robot end effector;
5. obs['target_object_pose']: the pose of the target object
6. obs['goal_pose']: the goal pose for the target object to reach
7. obs['distance_to_target']: the distance between the end effector and the target object.
8. obs['distance_to_goal']: the distance between the target object and the goal pose
9. obs['in_contact']: a boolean value if the robot is in contact with a target object if there is
10. obs['in_collision']: a boolean value if the robot is in collision with the environment
11. obs['action]: the actions robot applied in the last step

Note, you can check 
(1) binary grasped status between the gripper and the target object with grasped = env.robot.is_graspping(env.target_object)
(2) binary collision status between the robot and the environment with collision_detected = env.detect_collision()

We plan to learn different manipulation skills in this environment.
The robot is supposed to learn a sensorimotor policy with a neural network using reinforcement learning. When given observation as input the policy outputs the end effector pose action. The goal of the policy is to learn optimal action decision-making for a given skill.
\end{lstlisting}

\subsubsection{Reward Prompt with Task Description}

We parse all input arguments except observation and action as the parameters to update automatically. The reward features are automatically parsed inside the get\_reward() function as scalar local variables. As the reward signal is a scalar value, the features that are most relevant for such reward calculation should be a scalar as well. For replay buffer storage, instead of storing the raw reward value, we store the reward features. For reward relabeling, we calculate the new reward with the saved reward features using updated parameters. 

\begin{lstlisting}[breakautoindent=true, breakindent=10pt, breaklines]
To start, please analyze for task <task description>.

Please organize your answer into three parts:
1) describe the skill. Break down the task into substeps of actions as dos that the robot needs to achieve step by step, then turn them into reward terms step by step. During this sequence of actions, what are positive actions that will prompt more effective explorations? Include them in reward terms, Meanwhile, what should be avoided? Design them into penalty terms. 
2) The total reward should be a weighted sum of all reward and penalty terms. For each term, analyze what observations mentioned above are related to this term.   
3) Based on the analysis from 2), translate the reward design into Python function get_reward().

Please only use features that are semantically directly related to this task and minimize additional considerations such as regularization.

Note:
- Please only use observations that are available in obs_vars and put all hyper-parameters in the input argument with the recommended value as default input. 
- Do not assume and use functions that are not provided. 
- Do include any magic number in the reward function.
- All reward or penalty terms in the get_reward function should end with "_reward" or "_penalty". 
- In designing the get_reward() function, make sure you process the most related feature first and assign it to a variable with an intuitive name before calculating the reward. Otherwise, avoid naming the variables.
- If there are any hyper-parameters, put them as the input argument of the function with a recommended default value. Do not use any magic numbers in the function.

\end{lstlisting}

\subsection{Feedback and Ranking Prompt}
\begin{lstlisting}[breakautoindent=true, breakindent=5pt, breaklines]
The reward function you provided is:
    <get_reward()>

Given execution observation for
  - data sample 0: <feature_name> = <feature_value>, ...
  - data sample 1: <feature_name> = <feature_value>, ...
  - data sample 2: <feature_name> = <feature_value>, ...
  - data sample 3: <feature_name> = <feature_value>, ...
  - data sample 4: <feature_name> = <feature_value>, ...

First, summarize how many steps are there in the task.
Then go through the data samples one by one and identify which stage the execution is at.  Put objects at the same step into one cluster and list all the clusters. Note one sample should only belong to one cluster. Then rank samples from better to worse within a cluster for all clusters. Lastly, concatenate the ranking in a list by always putting the cluster at a later step in front.
Make sure the last line of the reply contains and only contains the final list.

Example of opening drawer task:
 - data 1, 4 are in the reaching stage, where 1 is closer than 4. The ranking for this cluster is [1, 4].
 - data 2, 5 are in the pulling stage, where 5 are pulled more than 2. The ranking for this cluster is [5, 2]
Pulling is at a later stage than reaching. The final result is:
[5, 2, 1, 4]
\end{lstlisting}

\subsection{Parameter Reflection Prompt}
\begin{lstlisting}[breakautoindent=true, breakindent=5pt, breaklines]
Given the reward function for <task_description> is:
    <get_reward()>

<repeat execution observation>

Go through each data sample and check if it succeeds in executing <skill description>. What action will encourage the current behaviour to be more likely to successfully execute open cabinet drawer? or if this is a multi-step task, which stages the current behaviour is at? what reward or penalty will prompt the current behaviour to produce meaningful exploration that contains the action of the next stage?

After going through all samples, count the times that each relevant reward or penalty term is mentioned.  Provide your chain of thought in plain text.

Lastly, output the identified hyper-parameter that is likely to prompt success or to the next stage behaviour as a dictionary.  The key is the hyper-parameter name and the value is the recommended new value. Comment behind each to indicate if the value is suggested to increase or decrease. Do not output anything after this.

For example:
Result:
{'param_a': 1.0, 'param_b': 1.0}
\end{lstlisting}

\subsection{Prompt to Generate Observation Dictionary Automatically for Isaac Gym Tasks}
As the original observations available are nicely structured and named in \texttt{compute\_observation()} in Isaac Gym tasks, we leverage LLM itself to automatically generate the observation dictionary within this function that later will be used for reward proposing and calculation. 

\begin{lstlisting}[breakautoindent=true, breakindent=5pt, breaklines]
You are an expert in robot manipulation. Now the task is to <task description>. Provided this is the compute_observations function which are the observations you can get:

<compute_observation()>

Please append a snippet of code which is supposed to be put at the last step of the compute_observation() function that aggregates all observations into a dictionary with the variable name "obs_vars", where the key is the observation name and value is the corresponding value.
\end{lstlisting}

\subsection{Task Description}

The task description used in the prompt template for the 9 evaluation tasks are listed below in table \ref{task_description}:
\begin{table}[ht]
    \begin{center}
    \begin{tabularx}{0.9\textwidth}{
    >{\raggedright\arraybackslash}X 
    >{\raggedright\arraybackslash}X }
     \hline
     \textit{Task} & \textit{Task Description} \\ 
     \hline
     \textit{Pick Cube} & pick cube object and transport to the target position \\ 
     \hline
     \textit{Pick Ycb} & pick ycb object and transport to the target position \\
     \hline
     \textit{Peg Insertion} & insert peg into the side hole \\
     \hline
     \textit{Open Cabinet Drawer} & open cabinet drawer as much as possible \\
     \hline
     \textit{Open Cabinet Door} & open cabinet door as much as possible\\
     \hline
     \textit{Push Chair} & push a swivel chair to a target 2D location on the ground\\
     \hline
     \textit{Franka Cabinet} & open cabinet drawer as much as possible \\
     \hline
     \textit{Shadow Hand Open Door Outward} & push the door as much as possible  with left and right panels outwards with two hands \\
     \hline
     \textit{Shadow Hand Scissor} & open the scissor as much as possible with two hands \\
    \hline
    \end{tabularx}
    \caption{Task descriptions for 9 tasks evaluated in the paper.}
    \label{task_description}
    \end{center}
\end{table}

\subsection{Example Reward Functions}
\subsubsection{Open Drawer}
\begin{lstlisting}[breakautoindent=true, breakindent=5pt, breaklines]
def get_reward(obs, alignment_weight=1.0, approach_weight=1.0, grasp_weight=1.0, pull_weight=1.0, collision_penalty_weight=1.0, non_progress_penalty_weight=1.0, distance_penalty_weight=0.1):

    # Extract relevant features from observations
    distance_to_handle = np.linalg.norm(obs['distance_to_target'])
    distance_to_goal = obs['distance_to_goal'][0]

    # Check for contact and collision
    contacted = env.robot.gripper_in_contact(env.target_object)
    collision_detected = env.detect_collision()

    alignment_reward = - alignment_weight * distance_to_handle

    # Approach reward: Encouraging the EE to get closer to the handle
    # Higher when the distance is smaller, so we take the negative distance
    approach_reward = - approach_weight * distance_to_handle

    # Grasp reward: Binary reward for making contact with the handle
    grasp_reward = grasp_weight * 1.0 if contacted else 0.0

    # Pull reward: Encouraging the EE to get the handle closer to the goal position
    # Higher when the distance is smaller, so we take the negative distance
    pull_reward = -pull_weight * distance_to_goal

    # Collision penalty: Penalizing any collision detected during interaction
    collision_penalty = -collision_penalty_weight * 1.0 if collision_detected else 0.0

    # Non-progress penalty: If there's no contact and no reduction in distance to the handle,
    # penalize to encourage the EE to move towards the handle
    non_progress_penalty = -non_progress_penalty_weight * 1.0 if not contacted and distance_to_handle > 0.02 else 0.0  # Threshold of 2cm

    # Excessive distance penalty: Discouraging the EE from being too far from the handle
    excessive_distance_penalty = - distance_penalty_weight * distance_to_handle if distance_to_handle > 0.15 else 0.0  # Threshold of 15cm

    # Combine rewards and penalties into total reward
    total_reward = alignment_reward + approach_reward + grasp_reward + pull_reward + collision_penalty + non_progress_penalty + excessive_distance_penalty

    return total_reward
\end{lstlisting}

\subsubsection{Push Chair}
\begin{lstlisting}[breakautoindent=true, breakindent=5pt, breaklines]
def get_reward(obs, approach_weight=1.0, movement_weight=1.0, collision_penalty_weight=1.0):

    # Reward for minimizing the distance between the robot gripper and the chair
    gripper_to_chair_dist = np.linalg.norm(obs['distance_to_target'])
    approach_reward = -gripper_to_chair_dist  # Negative value: smaller distance is better

    # Reward for moving the chair towards the target position
    # Assuming the target position is part of the environment's state
    chair_to_target_dist = np.linalg.norm(obs['distance_to_goal'][:2])
    movement_reward = -chair_to_target_dist  # Negative value: smaller distance is better

    # Penalty for collisions
    collision_detected = env.detect_collision()
    collision_penalty = -1.0 if collision_detected else 0.0

    # Calculate total reward
    total_reward = approach_weight * approach_reward + movement_weight * movement_reward + collision_penalty_weight * collision_penalty

    return total_reward
\end{lstlisting}

\subsubsection{Shadow Hand Open Scissor}
\begin{lstlisting}[breakautoindent=true, breakindent=5pt, breaklines]
def get_reward(obs, handle_separation_weight=1.0, hand_on_handle_weight=1.0, stability_weight=0.1):

    # Extract relevant features from obs
    scissors_right_handle_pos = obs['scissors_right_handle_pos']
    scissors_left_handle_pos = obs['scissors_left_handle_pos']

    left_hand_pos = obs['left_hand_pos']
    right_hand_pos = obs['right_hand_pos']

    object_linvel = obs['object_linvel']
    object_angvel = obs['object_angvel']

    # Initially, the handles are next to each other, we want to maximize the distance between the scissor handles
    handles_distance = torch.norm(scissors_right_handle_pos - scissors_left_handle_pos, p=2, dim=1)

    # Handle Separation Reward (encourage the hands to move the scissor handles apart)
    handle_separation_reward = handle_separation_weight * handles_distance

    # Hand on Handle Reward (encourage the hands to maintain contact with the scissor handles)
    left_hand_on_handle_distance = torch.norm(left_hand_pos - scissors_left_handle_pos, p=2, dim=1)
    right_hand_on_handle_distance = torch.norm(right_hand_pos - scissors_right_handle_pos, p=2, dim=1)
    hand_on_handle_reward = - hand_on_handle_weight * (left_hand_on_handle_distance + right_hand_on_handle_distance)

    # Stability Penalty (minimize the linear and angular velocity of the scissors to ensure smooth opening)
    linear_velocity_magnitude = torch.norm(object_linvel, p=2, dim=1)
    angular_velocity_magnitude = torch.norm(object_angvel, p=2, dim=1)
    stability_penalty = - stability_weight * (linear_velocity_magnitude + angular_velocity_magnitude)

    # Combine rewards and penalties
    total_reward = handle_separation_reward + hand_on_handle_reward + stability_penalty

    return total_reward
\end{lstlisting}

\subsection{Example Ranking from LLM}
\begin{lstlisting}[breakautoindent=true, breakindent=5pt, breaklines]
<Feedback>
|\color{gray}\textbf{Given execution observation for}|
  - data sample 0: distance_to_peg = 0.3771, alignment_error_ee_to_peg = 0.3695, grasped = False, collision_detected = True, peg_to_hole_distance = 0.1237, alignment_error_peg_to_hole = 0.0197.
  - data sample 1: distance_to_peg = 0.2108, alignment_error_ee_to_peg = 0.3156, grasped = False, collision_detected = False, peg_to_hole_distance = 0.12247, alignment_error_peg_to_hole = 0.0239.
  - data sample 2: distance_to_peg = 0.2523, alignment_error_ee_to_peg = 0.0064, grasped = False, collision_detected = False, peg_to_hole_distance = None, alignment_error_peg_to_hole = None.
  - data sample 3: distance_to_peg = 0.0914, alignment_error_ee_to_peg = 0.0174, grasped = True, collision_detected = False, peg_to_hole_distance = 0.3165, alignment_error_peg_to_hole = 0.0303.
  - data sample 4: distance_to_peg = 0.0474, alignment_error_ee_to_peg = 0.0139, grasped = True, collision_detected = False, peg_to_hole_distance = 0.2201, alignment_error_peg_to_hole = 0.0135.
  - data sample 5: distance_to_peg = 0.0334, alignment_error_ee_to_peg = 0.0106, grasped = True, collision_detected = False, peg_to_hole_distance = 0.1032, alignment_error_peg_to_hole = 0.0190.

<ChatGPT>
The task involves the following steps:

|\color{red}\textbf{Approaching the Peg}|: Reduce the distance to the peg and align the end effector with it.
|\color{red}\textbf{Grasping the Peg}|: Successfully grasp the peg.
|\color{red}\textbf{Transporting Peg to Hole}|: Move the peg towards the hole while maintaining alignment.
|\color{red}\textbf{Aligning Peg with Hole}|: Ensure the peg is properly aligned with the hole before insertion.
Now, let's analyze the data samples and categorize them based on the steps:

Data Sample Analysis

|\color{red}\textbf{Approaching the Peg Stage}|
Sample 0: Collision occurred, failed to grasp.
Sample 1: No collision, but failed to grasp. Better alignment than sample 0.
Sample 2: No collision, best alignment, but failed to grasp.

|\color{teal}\textbf{Approaching Cluster Ranking: [2, 1, 0]}|

|\color{red}\textbf{Grasping the Peg Stage}| (None)

|\color{red}\textbf{Transporting Peg to Hole Stage}| 
Sample 3: Grasped the peg, but the peg is far from the hole and alignment is not perfect.
Sample 4: Grasped, closer to the hole and better alignment than sample 3.
Sample 5: Grasped, closest to the hole and better alignment than sample 4.
|\color{teal}\textbf{Transporting Cluster Ranking: [5, 4, 3]}|

|\color{red}\textbf{Aligning Peg with Hole Stage (None)}|

Final Result
Putting the clusters in order based on the steps in the task, with later stages first:

[5, 4, 3, 2, 1, 0]
\end{lstlisting}

\subsection{Ranking Failure Cases}
\subsubsection{Small Numerical Difference}
As mentioned in Section \ref{boltzman_model}, LLM can mis-rank especially when there are subtle numerical differences. In the example shown below, data sample 4 should be the closest reaching but is ranked as 4th while other samples are ranked in the right order:
\begin{lstlisting}[breakautoindent=true, breakindent=5pt, breaklines]
Given execution observation for
  - data sample 0: drawer_joint_value = 0.0000, distance_to_drawer = |\color{teal}\textbf{0.0589}|, alignment = 0.9999.
  - data sample 1: drawer_joint_value = 0.0000, distance_to_drawer = |\color{teal}\textbf{0.0527}|, alignment = 1.0000.
  - data sample 2: drawer_joint_value = 0.0000, distance_to_drawer = |\color{teal}\textbf{0.0205}|, alignment = 1.0000.
  - data sample 3: drawer_joint_value = 0.0000, distance_to_drawer = |\color{teal}\textbf{0.2089}|, alignment = 0.9851.
  - data sample 4: drawer_joint_value = 0.0000, distance_to_drawer = |\color{red}\textbf{0.0168}|, alignment = 1.0000.
  
<analysis skipped>
All data samples are in the reaching stage, the ranking for the reaching stage cluster is [2, 1, 0, |\color{red}\textbf{4}|, 3] 
(|\color{red}\textbf{from best to worst}| distance_to_drawer).

\end{lstlisting}

\subsubsection{Ambiguous Task Description}
In general, to generate a ranking that is useful to guide the task policy, it is found that LLM requires the task description to be explicit about the task goal. When given an abstract description of a task such as "Pull a drawer", humans naturally associate with experience unspoken here, as a continuous change in the drawer prismatic joint and the more it is pulled the better. However, LLM tends to interpret words more literally with less such association. In the example shown below, LLM interpreted the task with the description "open the cabinet drawer" as a binary status goal. In the ranking of opened drawer cases, LLM seeks better orientation alignment between the end effector and opened drawer for higher ranking instead of the pulled length. Adding a goal such as "pull the cabinet drawer as much as possible" helps to remove such ambiguity, but in general, it hints at the requirement to carefully design the task description that no ambiguity presents, which can be less obvious for humans.

\begin{lstlisting}[breakautoindent=true, breakindent=5pt, breaklines]

Given execution observation for
  - data sample 0: drawer_joint_value = 0.0088, distance_to_drawer = 0.0988, alignment = 0.9998.
  - data sample 1: drawer_joint_value = 0.0510, distance_to_drawer = 0.1748, alignment = 0.9866.
  ...
  - data sample 4: drawer_joint_value = 0.0651, distance_to_drawer = 0.6472, alignment = 0.7687.

<analysis skipped>
Now we can cluster the data samples:
- Pulling stage: [0, 1, 4]

Within each cluster, we rank the samples from better to worse:
- Pulling stage: [0, 1, 4] (0 has the best alignment and smallest distance, 1 has the next best alignment and distance, 4 has the worst alignment and largest distance)
\end{lstlisting}

\subsection{Example Parameter Reflection from  LLM}
\begin{lstlisting}[breakautoindent=true, breakindent=5pt, breaklines]
Data sample 0:
- The robot has successfully grasped the peg and is moving towards the hole. The distance to the peg and the alignment error between the end effector and the peg are both low, indicating that the robot is holding the peg correctly. The distance to the hole and the alignment error between the peg and the hole are relatively high, suggesting that the robot is in the process of moving the peg towards the hole. No collision is detected. The robot should be encouraged to continue reducing the distance to the hole and aligning the peg with the hole.

Data sample 1:
- The robot is still holding the peg and has moved closer to the hole, as indicated by the reduced distance to the hole. The alignment error between the end effector and the peg has increased slightly, but the alignment error between the peg and the hole has also increased. This suggests that while the robot is moving closer to the hole, it may not be aligning the peg correctly. No collision is detected. The robot should be encouraged to improve the alignment between the peg and the hole while continuing to reduce the distance to the hole.

Data sample 2:
- The robot continues to hold the peg and move closer to the hole. Both the distance to the peg and the alignment error between the end effector and the peg have decreased, indicating that the robot is holding the peg more correctly. The distance to the hole and the alignment error between the peg and the hole have also decreased, suggesting that the robot is aligning the peg with the hole more accurately. No collision is detected. The robot should be encouraged to continue this behavior.

Data sample 3:
- The robot is still holding the peg, but the distance to the hole has increased, suggesting that the robot may have moved away from the hole. The alignment error between the end effector and the peg has slightly increased, but the alignment error between the peg and the hole has decreased significantly. This suggests that while the robot may have moved away from the hole, it has improved the alignment of the peg with the hole. No collision is detected. The robot should be encouraged to move closer to the hole while maintaining the alignment of the peg with the hole.

Data sample 4:
- The robot continues to hold the peg, but the distance to the hole has significantly increased, suggesting that the robot has moved further away from the hole. The alignment error between the end effector and the peg has also increased, indicating that the robot may not be holding the peg correctly. However, the alignment error between the peg and the hole has decreased significantly, suggesting that the robot has improved the alignment of the peg with the hole. No collision is detected. The robot should be encouraged to move closer to the hole and improve the way it holds the peg.

Count of relevant reward or penalty terms:
- approach_reward: 5 times
- ee_to_peg_alignment_reward: 5 times
- grasp_reward: 5 times
- lift_and_move_reward: 5 times
- peg_to_hole_alignment_reward: 5 times
- collision_penalty: 0 times

Result:
{'distance_to_peg_weight': 60.0, # |\color{teal}\textbf{increase}| to encourage the robot to reduce the distance to the peg
 'alignment_ee_to_peg_weight': 80.0, # |\color{teal}\textbf{increase}| to encourage the robot to improve the alignment between the end effector and the peg
 'grasp_reward_value': 70.0, # |\color{red}\textbf{decrease}| as the robot is already successfully grasping the peg
 'distance_to_hole_weight': 90.0, # i|\color{teal}\textbf{ncrease}| to encourage the robot to reduce the distance to the hole
 'alignment_peg_to_hole_weight': 75.0, # |\color{teal}\textbf{increase}| to encourage the robot to improve the alignment between the peg and the hole
 'collision_penalty_value': -20.0} # # |\color{blue}\textbf{no change}| as no collision is detected

\end{lstlisting}

\subsection{Visualization for Alignment over Iterations}

\begin{figure*}[ht]
\centering
\begin{subfigure}
  \centering
  \includegraphics[width=0.4\linewidth]{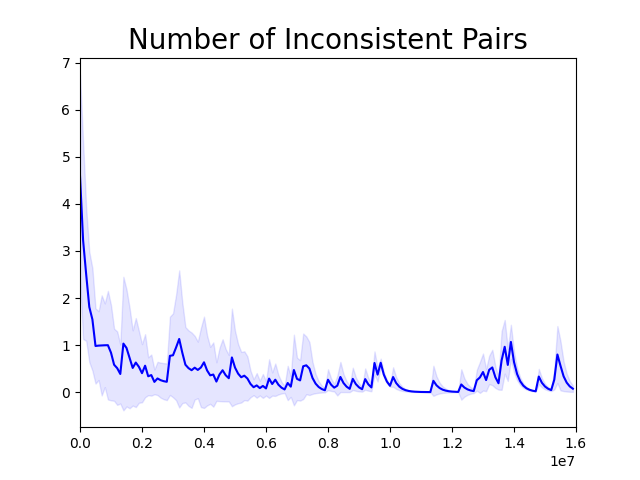}
  \label{fig:n_align_vs_iter}
\end{subfigure}
\begin{subfigure}
  \centering
  \includegraphics[width=0.4\linewidth]{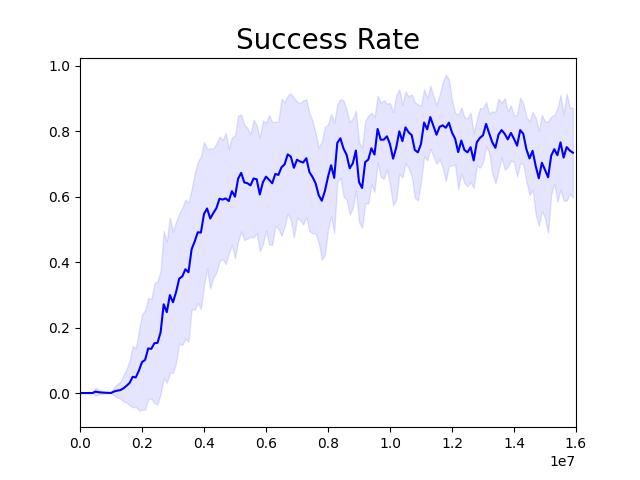}
  \label{fig:sr_vs_iter}
\end{subfigure}
\begin{subfigure}
  \centering
  \includegraphics[width=0.4\linewidth]{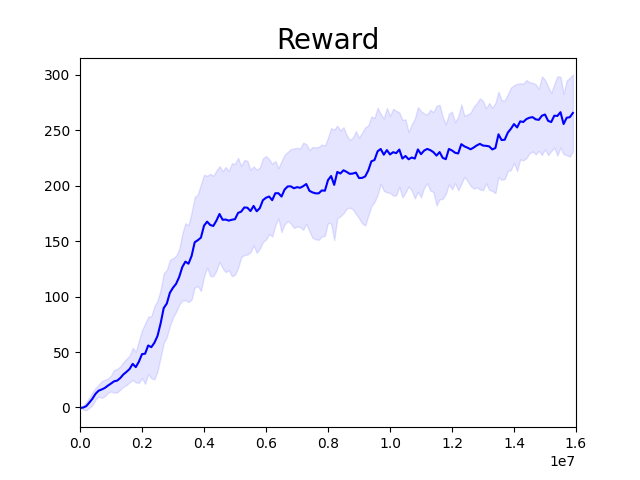}
  \label{fig:updated_reward_vs_iter.png}
\end{subfigure}
\begin{subfigure}
  \centering
  \includegraphics[width=0.4\linewidth]{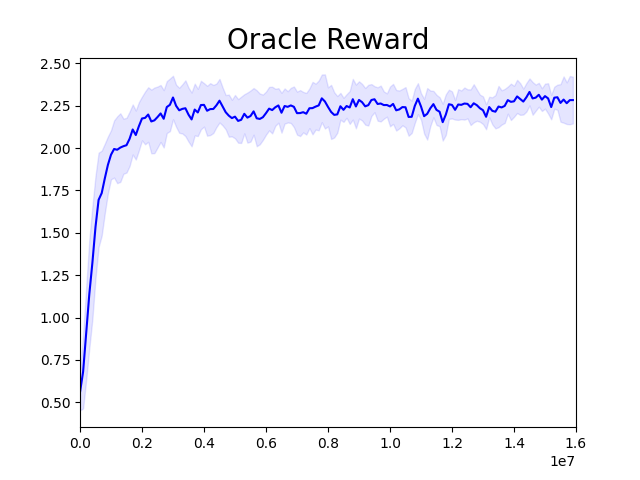}
  \label{fig:oracle_reward_vs_iter.png}
\end{subfigure}
\caption{The number of inconsistencies, success rate, self-alignment updated reward and the corresponding oracle reward with training steps for the pick cube task.}
\label{align_vis}
\end{figure*}
To better understand how the alignment evolves with the learning process, we plot the number of inconsistent pairs, success rate, self-alignment updated reward and the corresponding original oracle reward with training steps over five seeds in Fig \ref{align_vis}. It is observed that in the early stage, the inconsistency in general decreases quickly but also spikes with new sub-step behavior picked up. It gradually decrease to around 0 and maintain here.

\subsection{Visualization for Reward Update Triggers}

\begin{figure*}[ht]
\centering
\begin{subfigure}
  \centering
  \includegraphics[width=0.31\linewidth]{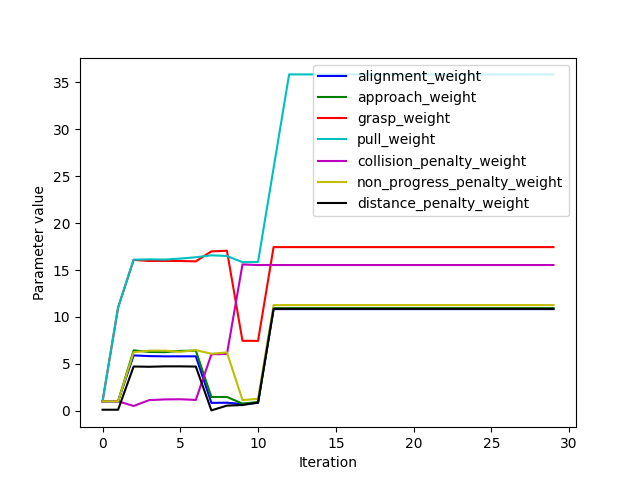}
  \label{fig:open_draw_wu}
\end{subfigure}
\begin{subfigure}
  \centering
  \includegraphics[width=0.31\linewidth]{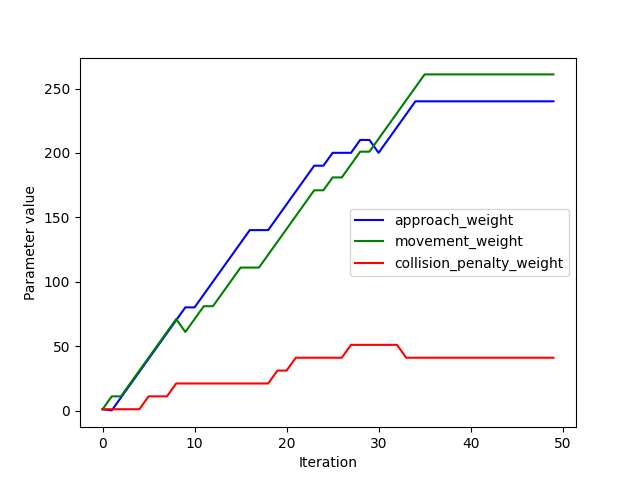}
  \label{fig:push_chair_wu}
\end{subfigure}
\begin{subfigure}
  \centering
  \includegraphics[width=0.31\linewidth]{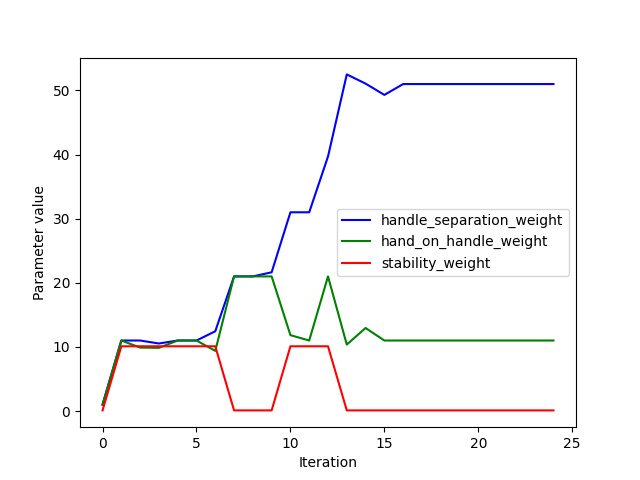}
  \label{fig:open_scissor.png}
\end{subfigure}
\caption{Visualization for weight update over iterations (1) Open Drawer, (2) Push Chair; (3) Shadow Hand Open Scissor}
\label{weight_update_appendix}
\end{figure*}

\begin{figure*}[ht]
    \centering
    \includegraphics[width=0.75\textwidth]{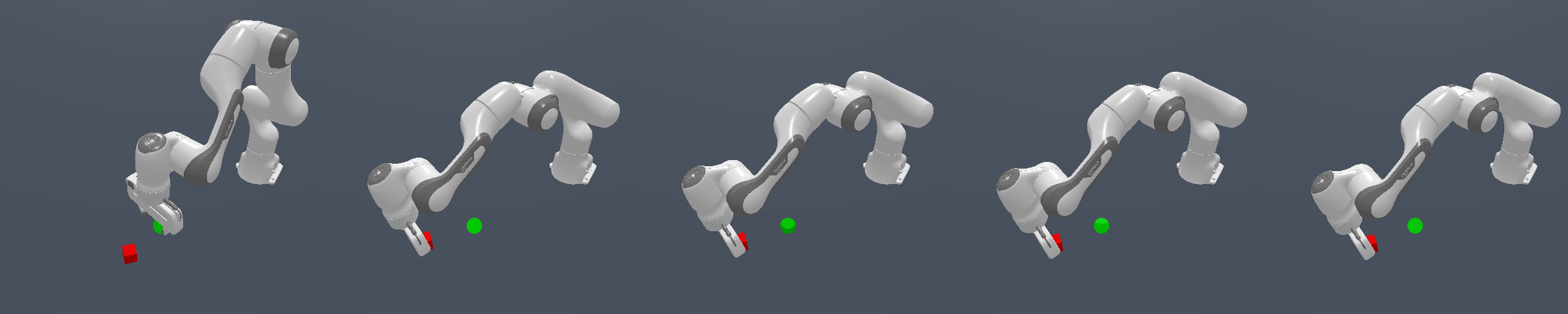}
    \includegraphics[width=0.75\textwidth]{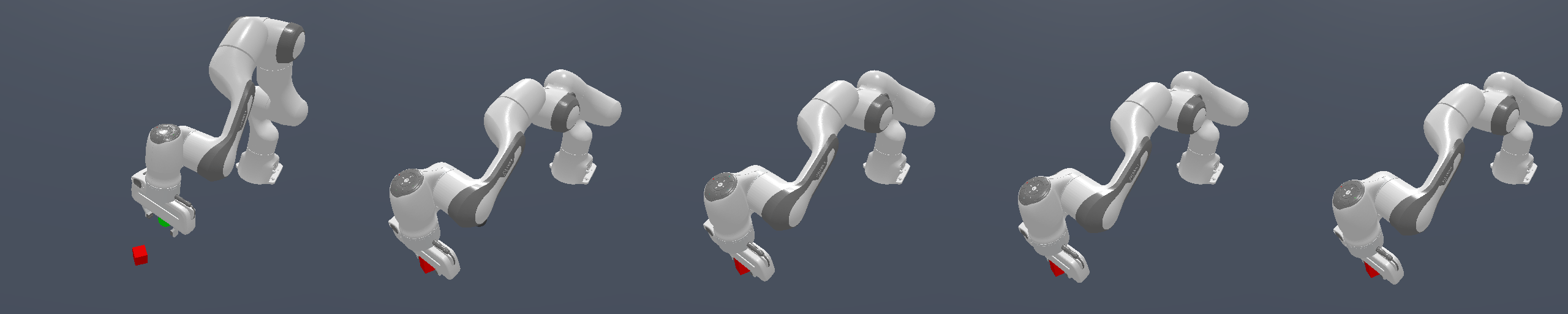}
    \includegraphics[width=0.75\textwidth]{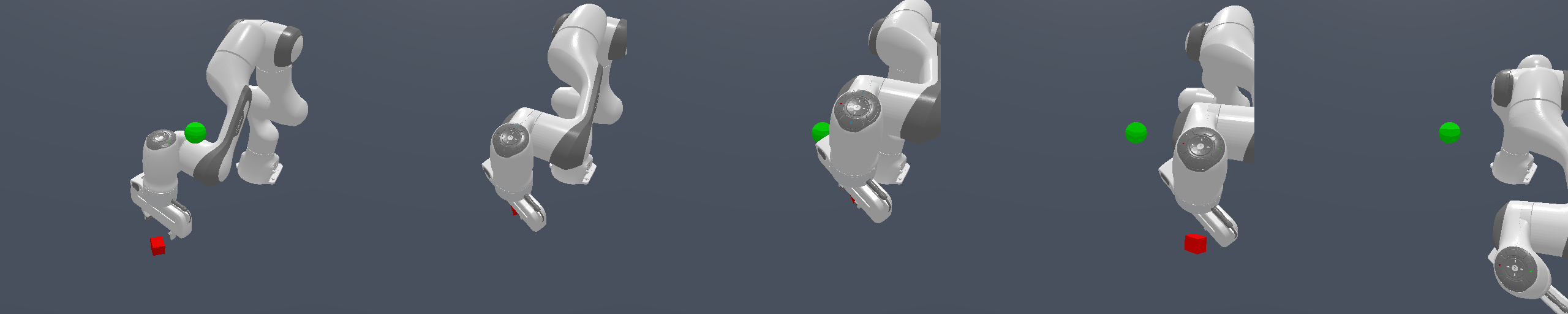}
    \includegraphics[width=0.75\textwidth]{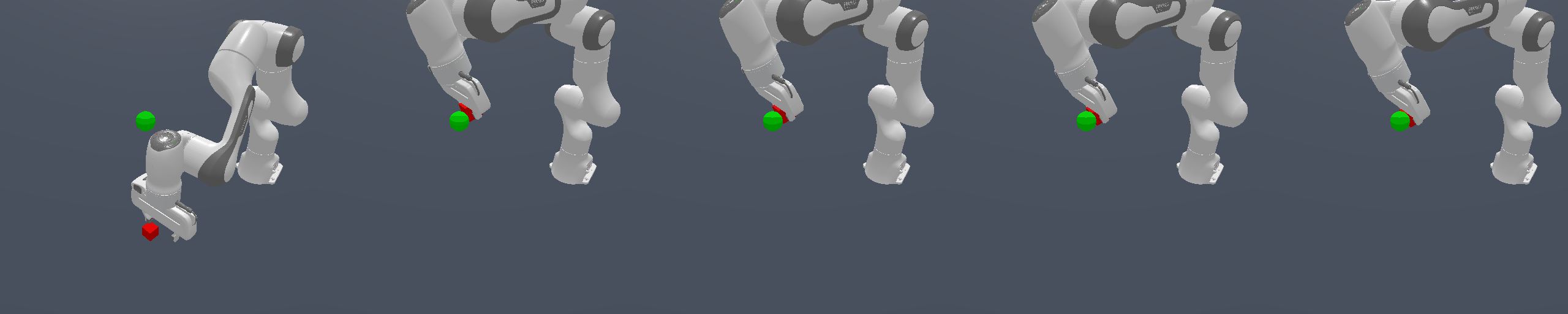}
\caption{Visualization of policy behaviour around iterations where major parameter changes as visualized in  Fig \ref{weight_curriculum} (a) were observed. They are (from top to bottom) at iteration 5, 12, 23, and 35. The weight updated is found to be interpretable and directly correlates to the policy behaviour.}
\label{fig:transition_visualization}
\end{figure*}

The weights for runs on tasks Open Drawer, Push Chair, Shadow Hand Open Scissors are plotted as shown in Fig \ref{weight_update_appendix}. Where we can observe weights-related parameters, especially for pick cube, open drawer and open scissor, weights tend to change transiting from approach, then grasp (if involved) then final goal reaching with the policy improving. We can also observe multiple cases where approaching/grasping weight decreases significantly to allow exploration for goal reaching stage.

In Fig \ref{fig:transition_visualization}. We further visualized the policy rollout at iteration 5, iteration 12, iteration 23 and iteration 35 as shown with major reward update pattern changes as shown in Fig \ref{weight_curriculum} (a). The weight updated is found to be interpretable and directly correlates to the policy behaviour:

\begin{itemize}
    \item  \textit{Iteration 5}: The model gradually starts to reach to object every time from random behavior, where it is observed the approaching weight starts to stabilize;
    \item \textit{Iteration 12}: The model starts to grasp the cube and move to random locations. Similarly, a slower increase in grasp weight is observed and goal-reaching and maintaining weight start to increase more rapidly;
    \item \textit{Iteration 23}: Goal reaching and maintaining weight slows down, and grasping and approaching weight starts to increase once again. From the rollout policy, the policy learns to pick up the cube and tries to reach the goal fast while the cube becomes unstable and slips out of hand where re-grasping happens.
    \item \textit{Iteration 35}: Goal reaching and maintaining weight keeps increasing while other parameters stay fixed. It is observed the robot can pick up the cube to go near the goal position although not exactly. The two weight terms stabilize in a few more iterations.
\end{itemize}


\end{document}